\documentclass{article}
\usepackage[utf8]{inputenc}
\usepackage{verbatim}
\usepackage{cprotect}
\usepackage{booktabs}
\usepackage{hepnames}
\usepackage{amsmath}
\usepackage{amsthm}
\usepackage{nomencl}
\makenomenclature
\usepackage[bookmarks=false,
 breaklinks=false,pdfborder={0 0 1},backref=section,colorlinks=false]
 {hyperref}
\hypersetup{
 hidelinks}

\makeatletter

\newcommand{\lyxmathsym}[1]{\ifmmode\begingroup\def\b@ld{bold}
  \text{\ifx\math@version\b@ld\bfseries\fi#1}\endgroup\else#1\fi}

\providecommand{\tabularnewline}{\\}

\theoremstyle{plain}
\newtheorem{thm}{\protect\theoremname}
\theoremstyle{definition}
\newtheorem{defn}[thm]{\protect\definitionname}
\theoremstyle{definition}
\newtheorem{example}[thm]{\protect\examplename}
\theoremstyle{plain}
\newtheorem{prop}[thm]{\protect\propositionname}
\theoremstyle{plain}
\newtheorem{cor}[thm]{\protect\corollaryname}
\theoremstyle{remark}
\newtheorem{rem}[thm]{\protect\remarkname}

\usepackage{kr}

\usepackage{times}
\usepackage{soul}
\usepackage{url}
\usepackage[small]{caption}
\usepackage{graphicx}
\usepackage{amsthm}
\usepackage{algorithm}
\usepackage{algorithmic}
\usepackage{latexsym}
\usepackage{booktabs}
\usepackage{enumitem}
\usepackage{color}

\usepackage{verbatim}
\usepackage{amsmath}
\usepackage{amssymb}

\usepackage{tikz}

\title{The Largest Tractable Subclass of Multi-Scale Allen Temporal Reasoning}

\author{%
    Quentin Cohen-Solal
    \affiliations
   LAMSADE, Université Paris-Dauphine, PSL, CNRS, Paris, France
    \emails
quentin.cohen-solal@dauphine.psl.eu    
}

\usepackage{cuted}

\makeatother

\providecommand{\corollaryname}{Corollary}
\providecommand{\definitionname}{Definition}
\providecommand{\examplename}{Example}
\providecommand{\propositionname}{Proposition}
\providecommand{\remarkname}{Remark}
\providecommand{\theoremname}{Theorem}

\begin{document}
\maketitle
\begin{abstract}
In this paper, we propose a novel model for temporal reasoning over
multi-scale intervals. We show that this model yields a complete reasoning
framework and exhibits other particularly interesting properties.
Furthermore, we identify the maximal tractable subalgebra for this
model, as well as for the other complete model, in the qualitative
scales context, which exhibits weaker properties but whose results
can be derived by transfer from our new model. 
\end{abstract}
\global\long\def\repr{\mathfrak{g}^{-1}}%
\global\long\def\gran{\mathfrak{g}}%
\global\long\def\liste#1#2{\left\{  #1\ |\ #2\right\}  }%

\global\long\def\Q{\mathbb{Q}}%
\global\long\def\pc{\mathrm{PC}}%
\global\long\def\pa{\mathrm{PA}}%
\global\long\def\gpa{\mathrm{GPA}}%

\global\long\def\J{\left\{  0,\ldots,m-1\right\}  }%

\global\long\def\Urgpc{\Q}%

\global\long\def\rmgpc{\mathrm{GPC}_{\Q}^{m}}%

\global\long\def\rmgic{\mathrm{GIC}_{\Q}^{m}}%

\global\long\def\rgpc{\mathrm{GPC}_{\Q}}%
\global\long\def\rgic{\mathrm{GIC}_{\Q}}%
\global\long\def\E{\mathtt{E}}%
\global\long\def\i{\mathfrak{i}}%
\global\long\def\S{\mathcal{S}}%
\global\long\def\upconv{\uparrow}%
\global\long\def\downconv{\downarrow}%
\global\long\def\inter{\mathbin{\cap}}%
\global\long\def\Repr{\mathfrak{G}}%
\global\long\def\Int{\mathcal{I}}%

\noindent\begin{minipage}[t]{1\columnwidth}%
\global\long\def\Cl#1{\mathfrak{C}\left(#1\right)}%

\global\long\def\id{\mathrm{id}}%

\global\long\def\conv#1#2{\mathop{\Rsh_{{#1}}^{{#2}}}}%

\global\long\def\cconv{\mathop{\Rsh}}%

\global\long\def\convDe#1#2#3{{_{{\scriptscriptstyle #3}}{\Rsh_{#1}^{#2}}}}%

\global\long\def\convinv#1#2{\mathop{{\mirrorRsh}_{{#1}}^{{#2}}}}%

\global\long\def\cconvinv{\mathop{\mirrorRsh}}%

\global\long\def\upgc{\mathop{\uparrow}}%

\global\long\def\downgc{\mathop{\downarrow}}%

\global\long\def\sinv#1{\overline{#1}}%

\global\long\def\scompl#1{#1^{\lhook}}%

\global\long\def\B{\mathcal{B}}%

\global\long\def\A{\mathcal{A}}%

\global\long\def\se{\mathrm{e}}%

\global\long\def\Base{\boldsymbol{\B}}%

\global\long\def\svide{\varnothing}%

\global\long\def\atome{\in}%

\global\long\def\comp{\circ}%

\global\long\def\wcomp{\diamond}%

\global\long\def\pcomp{\ast}%

\global\long\def\acomp{\varocircle}%

\global\long\def\ainter{\wedge}%

\global\long\def\aunion{\vee}%

\global\long\def\inv#1{\overline{#1}}%

\global\long\def\ainv#1{#1{}^{{\scriptscriptstyle \smallsmile}}}%

\global\long\def\winv#1{\widetilde{#1}}%

\global\long\def\acompl#1{\neg#1}%

\global\long\def\compl#1{#1^{\lhook}}%

\global\long\def\size#1{\mathrm{size}(#1)}%

\global\long\def\scomp{\varocircle}%

\global\long\def\sinter{\cap}%

\global\long\def\sunion{\cup}%

\global\long\def\sbigunion{\bigcup}%

\global\long\def\Ptime{\mathsf{P}}%

\global\long\def\PSPACE{\mathsf{PSPACE}}%

\global\long\def\NP{\mathsf{NP}}%

\global\long\def\T{\mathcal{T}}%

\global\long\def\U{\mathcal{U}}%

\global\long\def\I{\varphi}%

\global\long\def\topo{\mathfrak{T}}%

\global\long\def\clotureTopo#1{#1^{-}}%

\global\long\def\et{\,\wedge\,}%

\global\long\def\ou{\,\vee\,}%

\global\long\def\bigand{\qquad\wedge\qquad}%

\global\long\def\liste#1#2{\{#1\mid#2\}}%

\global\long\def\bigliste#1#2{\left\{  #1\mid#2\right\}  }%

\global\long\def\sep{\quad}%

\global\long\def\minisep{\ }%

\global\long\def\aeq{\mathrm{I}}%

\global\long\def\Id{\mathrm{Id}}%

\global\long\def\Ba{\top}%

\global\long\def\Br{\mathtt{B}}%

\global\long\def\Bw{\Re}%

\global\long\def\avide{\bot}%

\global\long\def\vide{\varnothing}%

\global\long\def\Aa{\mathcal{A}}%

\global\long\def\Au{\mathfrak{A}}%

\global\long\def\Ar{\mathtt{A}}%

\global\long\def\S{\mathcal{S}}%

\global\long\def\BaseR{\pmb{\Br}}%

\global\long\def\BaseA{\pmb{\Ba}}%

\global\long\def\BaseU{\mathfrak{B}}%

\global\long\def\PA{\mathrm{PA}}%

\global\long\def\IA{\mathrm{IA}}%

\global\long\def\PAs{\mathcal{S}_{\PA}}%

\global\long\def\PAc{\mathcal{C}_{\PA}}%

\global\long\def\IAs{\mathcal{S}_{\IA}}%

\global\long\def\IAc{\mathcal{C}_{\IA}}%

\global\long\def\CPA{\mathrm{CPA}}%

\global\long\def\ERA{\mathrm{ERA}}%

\global\long\def\INDU{\mathrm{INDU}}%

\global\long\def\QTC{\mathrm{QTC}}%

\global\long\def\MBRC{\mathrm{MBRC}}%

\global\long\def\DLR{\mathrm{DLR}}%

\global\long\def\hornDLR{\mathrm{hornDLR}}%

\global\long\def\RCC{\mathrm{RCC}}%

\global\long\def\RCA{\mathrm{RCA}}%

\global\long\def\RCCH{\RCC_{8}}%

\global\long\def\RCAH{\mathrm{\RCA_{8}}}%

\global\long\def\HH{\mathcal{\hat{H}}_{8}}%

\global\long\def\CH{\mathcal{C}_{8}}%

\global\long\def\QH{\mathcal{Q}_{8}}%

\global\long\def\RCAHs{\mathcal{S}_{\RCAH}}%

\global\long\def\RCAHc{\mathcal{C}_{\RCAH}}%

\global\long\def\RCCF{\RCC_{5}}%

\global\long\def\RCAF{\RCA_{5}}%

\global\long\def\Aseize{\mathrm{A}_{16}}%

\global\long\def\Acinq{\mathrm{A}_{5}}%

\global\long\def\PC{\mathrm{PC}}%

\global\long\def\IC{\mathrm{IC}}%

\global\long\def\QS{\mathrm{QS}}%

\global\long\def\CPC{\mathrm{CPC}}%

\global\long\def\CRC{\mathrm{CRC}}%

\global\long\def\CDC{\mathrm{CDC}}%

\global\long\def\preconv{\mathcal{P}}%

\global\long\def\IAp{\preconv_{\IA}}%

\global\long\def\CPAp{\preconv_{\CPA}}%

\global\long\def\Arp{\preconv_{\Ar}}%

\global\long\def\Arc{\mathcal{C}_{\Ar}}%

\global\long\def\nord{\mathrm{N}}%

\global\long\def\sud{\mathrm{S}}%

\global\long\def\ouest{\mathrm{W}}%

\global\long\def\est{\mathrm{E}}%

\global\long\def\nordouest{\nord\ouest}%

\global\long\def\nordest{\nord\est}%

\global\long\def\sudouest{\sud\ouest}%

\global\long\def\sudest{\sud\est}%

\global\long\def\notni{\reflectbox{\ensuremath{\notin}}}%

\global\long\def\connexion{\mathrm{C}}%

\global\long\def\p{\mathrm{P}}%

\global\long\def\pI{\overline{\mathrm{P}}}%

\global\long\def\o{\mathrm{O}}%

\global\long\def\dc{\mathrm{DC}}%

\global\long\def\ec{\mathrm{EC}}%

\global\long\def\po{\mathrm{PO}}%

\global\long\def\tpp{\mathrm{TPP}}%

\global\long\def\ntpp{\mathrm{NTPP}}%

\global\long\def\tppi{\mathrm{\overline{TPP}}}%

\global\long\def\ntppi{\mathrm{\overline{NTPP}}}%

\global\long\def\eq{\mathrm{EQ}}%

\global\long\def\pp{\mathrm{PP}}%

\global\long\def\ppi{\mathrm{\overline{PP}}}%

\global\long\def\contrainte#1#2#3{#1\mathrel{#2}#3}%

\global\long\def\E{\mathtt{E}}%

\global\long\def\CdeN{\mathtt{C}}%

\global\long\def\miniunion{{\scriptscriptstyle \cup}}%

\global\long\def\lsun#1{\{\mathord{#1}\}}%

\global\long\def\lsdeux#1#2{\{\mathord{#1},\ \mathord{#2}\}}%

\global\long\def\lstrois#1#2#3{\{\mathord{#1},\ \mathord{#2},\ \mathord{#3}\}}%

\global\long\def\lsquatre#1#2#3#4{\{\mathord{#1},\ \mathord{#2},\ \mathord{#3},\ \mathord{#4}\}}%

\global\long\def\lscinq#1#2#3#4#5{\{\mathord{#1},\ \mathord{#2},\ \mathord{#3},\ \mathord{#4},\ \mathord{#5}\}}%

\global\long\def\lssix#1#2#3#4#5#6{\{\mathord{#1},\ \mathord{#2},\ \mathord{#3},\ \mathord{#4},\ \mathord{#5},\ \mathord{#6}\}}%

\global\long\def\lssept#1#2#3#4#5#6#7{\{\mathord{#1},\ \mathord{#2},\ \mathord{#3},\ \mathord{#4},\ \mathord{#5},\ \mathord{#6},\ \mathord{#7}\}}%

\global\long\def\lshuit#1#2#3#4#5#6#7#8{\{\mathord{#1},\ \mathord{#2},\ \mathord{#3},\ \mathord{#4},\ \mathord{#5},\ \mathord{#6},\ \mathord{#7},\ \mathord{#8}\}}%

\global\long\def\lsneuf#1#2#3#4#5#6#7#8#9{\{\mathord{#1},\ \mathord{#2},\ \mathord{#3},\ \mathord{#4},\ \mathord{#5},\ \mathord{#6},\ \mathord{#7},\ \mathord{#8},\ \mathord{#9}\}}%
\global\long\def\un#1{\{\mathord{#1}\}}%

\global\long\def\deux#1#2{\{\mathord{#1},\mathord{#2}\}}%

\global\long\def\trois#1#2#3{\{\mathord{#1},\mathord{#2},\mathord{#3}\}}%

\global\long\def\quatre#1#2#3#4{\{\mathord{#1},\mathord{#2},\mathord{#3},\mathord{#4}\}}%

\global\long\def\cinq#1#2#3#4#5{\{\mathord{#1},\mathord{#2},\mathord{#3},\mathord{#4},\mathord{#5}\}}%

\global\long\def\six#1#2#3#4#5#6{\{\mathord{#1},\mathord{#2},\mathord{#3},\mathord{#4},\mathord{#5},\mathord{#6}\}}%

\global\long\def\sept#1#2#3#4#5#6#7{\{\mathord{#1},\mathord{#2},\mathord{#3},\mathord{#4},\mathord{#5},\mathord{#6},\mathord{#7}\}}%

\global\long\def\huit#1#2#3#4#5#6#7#8{\{\mathord{#1},\mathord{#2},\mathord{#3},\mathord{#4},\mathord{#5},\mathord{#6},\mathord{#7},\mathord{#8}\}}%

\global\long\def\neuf#1#2#3#4#5#6#7#8#9{\{\mathord{#1},\mathord{#2},\mathord{#3},\mathord{#4},\mathord{#5},\mathord{#6},\mathord{#7},\mathord{#8},\mathord{#9}\}}%

\global\long\def\F{\mathcal{F}}%

\global\long\def\STC{\mathrm{STC}}%

\global\long\def\hmax{h_{\mathrm{max}}}%

\global\long\def\hHH{h_{\HH}}%

\global\long\def\hQH{h_{\QH}}%

\global\long\def\hCH{h_{\CH}}%

\global\long\def\correctTable{\raisebox{3.5mm}{}}%

\noindent\begin{minipage}[t]{1\columnwidth}%
\global\long\def\scomp{\varocircle}%

\global\long\def\sinter{\cap}%

\global\long\def\sunion{\cup}%

\global\long\def\sbigunion{\bigcup}%

\global\long\def\sinv#1{\overline{#1}}%

\global\long\def\scompl#1{#1^{\lhook}}%

\global\long\def\B{\mathcal{B}}%

\global\long\def\A{\mathcal{A}}%

\global\long\def\se{\mathrm{e}}%

\global\long\def\Base{\boldsymbol{\B}}%

\global\long\def\svide{\varnothing}%

\global\long\def\atome{\in}%

\end{minipage}

\noindent\begin{minipage}[t]{1\columnwidth}%
\global\long\def\arbo{\mathrm{A}}%

\global\long\def\inArbo{\overset{\arbo}{\rightarrow}}%

\global\long\def\inA{\rightarrow}%

\global\long\def\inAinv{\leftarrow}%
\end{minipage}

\noindent\begin{minipage}[t]{1\columnwidth}%
\global\long\def\RCCS{\mathrm{\RCC_{7}}}%

\global\long\def\DRA{\mathcal{DRA}_{24}}%

\global\long\def\SPA{\mathrm{SPA}}%

\global\long\def\SIA{\mathrm{SIA}}%

\global\long\def\SPIA{\mathrm{SPIA}}%

\global\long\def\SPC{\mathrm{SPC}}%

\global\long\def\SIC{\mathrm{SIC}}%

\global\long\def\SPIC{\mathrm{SPIC}}%

\global\long\def\STA{\mathrm{STA}}%

\global\long\def\PIA{\mathrm{PIA}}%

\global\long\def\PIC{\mathrm{PIC}}%
\end{minipage}

\noindent\begin{minipage}[t]{1\columnwidth}%
\global\long\def\cun#1{(\mathord{#1})}%

\global\long\def\cdeux#1#2{(\mathord{#1},\mathord{#2})}%

\global\long\def\ctrois#1#2#3{(\mathord{#1},\mathord{#2},\mathord{#3})}%

\global\long\def\cquatre#1#2#3#4{(\mathord{#1},\mathord{#2},\mathord{#3},\mathord{#4})}%

\global\long\def\ccinq#1#2#3#4#5{(\mathord{#1},\mathord{#2},\mathord{#3},\mathord{#4},\mathord{#5})}%

\global\long\def\csix#1#2#3#4#5#6{(\mathord{#1},\mathord{#2},\mathord{#3},\mathord{#4},\mathord{#5},\mathord{#6})}%

\global\long\def\csept#1#2#3#4#5#6#7{(\mathord{#1},\mathord{#2},\mathord{#3},\mathord{#4},\mathord{#5},\mathord{#6},\mathord{#7})}%

\global\long\def\chuit#1#2#3#4#5#6#7#8{(\mathord{#1},\mathord{#2},\mathord{#3},\mathord{#4},\mathord{#5},\mathord{#6},\mathord{#7},\mathord{#8})}%

\global\long\def\cneuf#1#2#3#4#5#6#7#8#9{(\mathord{#1},\mathord{#2},\mathord{#3},\mathord{#4},\mathord{#5},\mathord{#6},\mathord{#7},\mathord{#8},\mathord{#9})}%
\end{minipage}

\global\long\def\prodDyna{\otimes}%
\nomenclature[ST Ensemble]{$\prodDyna$}{$\Cvierge \prodDyna \Cvierge'$ est l'ensemble des évolutions de couples d'entités correspondant à l'ensemble des couples d'évolutions $\Cvierge \times \Cvierge'$ de deux dynamiques $\Cvierge$ et $\Cvierge'$ (déf. \vref{def:produit-dynamique}).}

\global\long\def\unpack#1{#1^{\star}}%
\nomenclature[ST Ensemble]{$\unpack{\Cvierge}$}{Ensemble des couples d'évolutions correspondant à l'ensemble des évolutions de couples $\Cvierge$ (déf. \vref{def:unpack}).}

\global\long\def\fcarac{\mathrm{c}}%

\global\long\def\voisinage{\mathrm{p}}%

\global\long\def\dominance{\mathrm{d}}%

\global\long\def\Fcarac#1#2{{#1}^{#2}}%
\nomenclature[ST Formalisme 0]{$\Fcarac{\F}{\fcarac}$}{Formalisme des relations de $\F$ sur la caractéristique $\fcarac$ des entités de $\U$ (déf. \vref{def:changement-semantique}).}

\global\long\def\interpCarac#1#2{#1^{#2}}%
\nomenclature[ST Fonction]{$\interpCarac{\I}{\fcarac}$}{Interprétation selon la caractéristique $\fcarac$ : $\interpCarac{\I}{\fcarac}(r)=\liste{(x,y)\in\U\times\U}{\contrainte{\fcarac(x)}{\I(r)}{\fcarac(y)}}$ (déf. \vref{def:changement-semantique}).}

\global\long\def\carac{\mathscr{C}}%

\global\long\def\mesure{\mathfrak{t}}%

\global\long\def\mesureLebesgue{\mathbf{\mathfrak{\boldsymbol{t}}}}%
\nomenclature[ST Fonction]{$\mesureLebesgue$}{Mesure de Lebesgue (voir la section  \vref{subsec:Definitions-mathematiques-continuite-taille}).}

\global\long\def\IQSF{\I^{\fcarac}}%

\global\long\def\Tv#1{{\mathrm{T}#1}^{\mathrm{n}}}%
\nomenclature[ST Formalisme 1]{$\Tv{\F}^{\Cvierge}$}{Formalisme $\F$ temporalisé sans relation intermédiaire par la dynamique $\Cvierge$ (déf. \vref{def:TFv}).}

\global\long\def\Tvf#1{{\mathrm{T}#1}^{\voisinage'}}%

\global\long\def\Td#1{{\mathrm{T}#1}^{\mathrm{d}}}%
\nomenclature[ST Formalisme 1]{$\Td{\F}^{\Cvierge}$}{Formalisme $\F$ temporalisé sur intervalles  par la dynamique $\Cvierge$ (déf. \vref{def:TFd}).}

\global\long\def\Tt#1{{\mathrm{T}#1}^{\mesure}}%
\nomenclature[ST Formalisme 2]{$\Tt{\RCCH}$}{Calcul des régions connectés et mesurables temporalisé avec relations de taille (il se note également $\Tc{\RCCm}{\PC}{\mesureLebesgue}{}$ ; déf. \vref{def:TMS}).}\nomenclature[ST Formalisme 2]{$\Tc{\F}{\tilde{\F}}{\fcarac}{\Cvierge}$}{Formalisme $\F$ temporalisé par la dynamique $\Cvierge$ avec relations de $\tilde{F}$ sur la caractéristique $\fcarac$ (déf. \vref{def:TMS}).}

\global\long\def\Tdt#1{{\mathrm{T}#1}^{\dominance,\mesure}}%
\nomenclature[ST Formalisme 3]{$\Tdc{\F}{\tilde{\F}}{\fcarac}{\Cvierge}$}{Formalisme $\F$ temporalisé sur intervalles par la dynamique $\Cvierge$ avec relations de $\tilde{F}$ sur la caractéristique $\fcarac$ (déf. \vref{def:Fdc}).}

\global\long\def\Tvt#1{{\mathrm{T}#1}^{\voisinage,\mesure}}%
\nomenclature[ST Formalisme 3]{$\Tvc{\F}{\tilde{\F}}{\fcarac}{\Cvierge}$}{Formalisme $\F$ temporalisé sans relation intermédiaire par la dynamique $\Cvierge$ avec relations de $\tilde{F}$ sur la caractéristique $\fcarac$ (déf. \vref{def:Fvc}).}

\nomenclature[ST Opérateur]{$\conv{\F}{\Fcarac{\tilde{F}}{\fcarac}}$}{Projection de l'algèbre $\A$ vers l'algèbre $\tilde{\A}$  de l'intégration lâche de $\F$ et $\Fcarac{\tilde{\F}}{\fcarac}$ (déf. \vref{def:projection_integration_F_QS}).}

\global\long\def\Tc#1#2#3#4{{\mathrm{T}#1}^{#4}+\Fcarac{#2}{#3}}%

\global\long\def\Tdc#1#2#3#4{{\Td{#1}}^{#4}+\Fcarac{#2}{#3}}%

\global\long\def\Tvc#1#2#3#4{{\Tv{#1}}^{#4}+\Fcarac{#2}{#3}}%

\global\long\def\Te#1{{\mathrm{T}#1}_{\ast}}%

\global\long\def\Tec#1#2#3#4{{\Te{#1}}^{#4}+\Fcarac{#2}{#3}}%

\global\long\def\convFT#1{\conv{#1}{\mesure}}%

\global\long\def\convTF#1{\conv{\mesure}{#1}}%

\global\long\def\SRI#1#2{\mathrm{SRI}_{#1}^{#2}}%
\nomenclature[ST Relation]{$\SRI{[\tau,\tau']}{\F}$}{Dynamique de $\F$ sans relation intermédiaire durant l'intervalle $[\tau,\tau']$ : \\$\bigliste{(x,y)}{\exists t\in[\tau,\tau']\minisep\exists b,b'\in\B\sep(x,y)\in\interpTemporel b{[\tau,t[}{\Cvierge}\cap\interpTemporel{b\sunion b'}t{\Cvierge}\cap\interpTemporel{b'}{]t,\tau']}{\Cvierge}}$ (déf. \vref{def:SRI}).}

\global\long\def\FSRI#1#2{\mathrm{FSRI}_{#1}^{#2}}%

\global\long\def\interpTemporel#1#2#3{\left\langle #1\right\rangle _{#2}^{#3}}%
\nomenclature[ST Relation]{$\interpTemporel rI{\Cvierge}$}{Relation sur $\Cvierge$, interprétation de $r\in\A$ durant l'intervalle $I$ :\\$\interpTemporel rI{\Cvierge}=\liste{(x,y)\in\Cvierge}{\exists b\in r\minisep\forall t\in I\sep\contrainte{x(t)}{\I(b)}{y(t)}}$ (déf. \vref{def:relations_temporalisees}).}\nomenclature[ST Relation]{$\interpTemporel rt{\Cvierge}$}{Relation sur $\Cvierge$, interprétation de $r\in\A$ à l'instant t :\\$\interpTemporel rt{\Cvierge}=\liste{(x,y)\in\Cvierge}{\contrainte{x(t)}{\I(r)}{y(t)}}$ (déf. \vref{def:relations_temporalisees}).}

\global\long\def\interpTemporelTaille#1#2#3{\left\langle #1\right\rangle _{#2}^{#3}}%

\global\long\def\TPA{\Tv{\PA}}%

\global\long\def\TPC{\Tv{\PC}}%

\global\long\def\postdomine#1{\dashv_{#1}}%

\global\long\def\predomine#1{\vdash_{#1}}%

\global\long\def\domine#1{\succ_{#1}}%

\global\long\def\estdomine#1{\prec_{#1}}%

\global\long\def\perturbe#1{\rightsquigarrow_{#1}}%

\global\long\def\voisine#1{\wr_{#1}}%

\nomenclature[ST Opérateur]{$\conv{\mathfrak{r}}{}$ }{Projection du graphe de transition $\mathfrak{r}$, avec $\mathfrak{r}\in\{\wr,\succ,\prec,\rightsquigarrow,\vdash,\dashv\}$ : $\conv{\mathfrak{r}}{}b=\sbigunion\liste{b'\in\Base}{\contrainte b{\mathfrak{r}}{b'}}$ (déf. \vref{def:projections_espace_modes_topo}).}

\global\long\def\cpostdomine#1#2{\dashv_{#1}^{#2}}%
\nomenclature[ST Relation transition 2]{$\cpostdomine{\F}{\Cvierge}$}{$b \cpostdomine{\F}{\Cvierge} b'$ si et seulement s'il existe $(x,y)\in\Cvierge$ et $\tau,\tau'\in I$ tels quel $\tau<\tau'$,  $(x(t),y(t))\in\I(b)$ pour tout $t\in]\tau,\tau'[$ et $(x(\tau'),y(\tau'))\in\I(b')$ (déf. \vref{def:modele_espace_modes_topo} et déf. \vref{def:espace-de-modes_de_FQ}).}

\global\long\def\cpredomine#1#2{\vdash_{#1}^{#2}}%
\nomenclature[ST Relation transition 1]{$\cpredomine{\F}{\Cvierge}$}{$b \cpredomine{\F}{\Cvierge} b'$ si et seulement s'il existe $(x,y)\in\Cvierge$ et $\tau,\tau'\in I$ tels quel $\tau<\tau'$, $(x(\tau),y(\tau))\in\I(b)$ et $(x(t),y(t))\in\I(b')$ pour tout $t\in]\tau,\tau'[$ (déf. \vref{def:modele_espace_modes_topo} et déf. \vref{def:espace-de-modes_de_FQ}).}

\global\long\def\cdomine#1#2{\succ_{#1}^{#2}}%
\nomenclature[ST Relation transition 3]{$\cdomine{\F}{\Cvierge}$}{$b\cdomine{\F}{\Cvierge} b'$ si et seulement si $b\cpredomine{\F}{\Cvierge} b'\ou b'\cpostdomine{\F}{\Cvierge} b$ (déf. \vref{def:graphe_voisinage_formel}, déf. \vref{def:modele_espace_modes_topo} et déf. \vref{def:espace-de-modes_de_FQ}).}

\global\long\def\cvoisine#1#2{\wr_{#1}^{#2}}%
\nomenclature[ST Relation transition 5]{$\cvoisine{\F}{\Cvierge}$}{$ b\cvoisine{\F}{\Cvierge} b'$ si et seulement si $b\cperturbe{\F}{\Cvierge} b'\ou b\cperturbe{\F}{\Cvierge} b'$ (déf. \vref{def:graphe_voisinage_formel}, déf. \vref{def:modele_espace_modes_topo} et déf. \vref{def:espace-de-modes_de_FQ}).}

\global\long\def\cperturbe#1#2{\rightsquigarrow_{#1}^{#2}}%
\nomenclature[ST Relation transition 4]{$\cperturbe{\F}{\Cvierge}$}{$ b\cperturbe{\F}{\Cvierge} b'$ si et seulement si $b\cpostdomine{\F}{\Cvierge} b'\ou b\cpredomine{\F}{\Cvierge} b'$ (déf. \vref{def:graphe_voisinage_formel}, déf. \vref{def:modele_espace_modes_topo} et déf. \vref{def:espace-de-modes_de_FQ}).}

\global\long\def\tpostdomine#1{\cpostdomine{#1}{\mesure}}%

\global\long\def\tpredomine#1{\cpredomine{#1}{\mesure}}%

\global\long\def\tdomine#1{\cdomine{#1}{\mesure}}%

\global\long\def\tvoisine#1{\cvoisine{#1}{\mesure}}%

\global\long\def\tperturbe#1{\cperturbe{#1}{\mesure}}%

\global\long\def\Cvierge{\mathcal{C}}%
\nomenclature[ST Ensemble]{$\Cvierge$}{Une dynamique (déf. \vref{def:Dynamique}).}\nomenclature[ST Ensemble]{$\sqrt{\Cvierge}$}{Ensemble des évolutions d'entités de la dynamique $\Cvierge$ : $\sqrt{\Cvierge}=\liste x{\exists y\minisep (x,y)\in\Cvierge}$ (déf. \vref{def:sqrt-dynamique}).}

\global\long\def\C#1#2{\Cvierge_{#1}^{#2}}%
\nomenclature[ST Ensemble 1]{$\C{\topo}{I}$}{Dynamique générale de $\topo$ sur un intervalle $I$ (déf. \vref{def:dynamique-generale}).}\nomenclature[ST Ensemble 2]{$\C{\F}{I}$}{Dynamique générale de $\F$ sur un intervalle $I$ (déf. \vref{def:dynamique_FQ-dyna_symetrique}).}

\global\long\def\Cc#1#2#3{\Cvierge_{#1,#3}^{#2}}%
\nomenclature[ST Ensemble 3]{$\Cc{\topo}{I}{\fcarac}$}{Dynamique de la préservation de la caractéristique $\fcarac$ des entités de $\topo$ sur un intervalle $I$ :\\$\liste{f\in\C{\topo}{I}}{\forall t,t'\in I\sep\fcarac(f(t))=\fcarac(f(t'))}$ (déf. \vref{def:dynamique_caracteristique}).}\nomenclature[ST Ensemble 4]{$\Cc{\F}{I}{\fcarac}$}{Dynamique de la préservation de la caractéristique $\fcarac$ de $\F$ sur un intervalle $I$ (déf. \vref{def:dynamique-FQ-caracteristique}).}

\global\long\def\Ct#1#2{\Cc{#1}{#2}{\mesure}}%

\global\long\def\reg#1{\mathfrak{R}_{#1}}%

\global\long\def\regc#1{\mathfrak{R}_{#1}^{\mathrm{c}}}%

\global\long\def\regcm#1{\mathfrak{R}_{#1}^{\mathrm{c,m}}}%

\global\long\def\dh#1#2{d_{\mathrm{H}}(#1,#2)}%

\global\long\def\P#1#2{\smash{P_{#1}^{#2}}}%

\global\long\def\pRecb#1#2{P_{#1}(#2)}%

\global\long\def\rank#1{\operatorname{rang}(#1)}%

\global\long\def\nbsub#1{\mathcal{N}_{#1}}%

\global\long\def\Urccmb{{\mathfrak{R}_{\mathrm{b}}^{\mathrm{m}}}}%

\global\long\def\Urccm{{\mathfrak{R}_{\mathrm{m}}}}%
\nomenclature[ST Ensemble]{$\Urccm$}{Ensemble des régions mesurables de $\mathbb{R}^{n}$ (déf. \vref{def:Rm}).}

\global\long\def\RCCmb{\RCCH^{\mathrm{b,m}}}%

\global\long\def\RCCm{{\RCC_{8,\mathrm{m}}}}%
\nomenclature[ST Formalisme 0]{$\RCCm$}{$\RCCH$ restreint aux régions mesurables (déf. \vref{def:RCCHm}).}

\global\long\def\interpRCCmb{\I_{\mathrm{b,m}}}%

\global\long\def\interpRCCm{\I_{\mathrm{m}}}%
\nomenclature[ST Fonction]{$\interpRCCm$}{Interprétation de $\RCCH$ restreinte aux régions mesurables (déf. \vref{def:RCCHm}).}

\global\long\def\c{\mathrm{C}}%
\global\long\def\P{\mathcal{P}}%
\global\long\def\Nrcc{\mathcal{N}}%
\global\long\def\NPrcc{\mathcal{NP}_{8}}%
\global\long\def\Prcc{\mathcal{P}_{8}}%

\global\long\def\icclo{\mathord{\scomp\sinter}\text{-closed}}%
\end{minipage}

\section{Introduction}

Qualitative reasoning is a core paradigm in artificial intelligence
for representing and reasoning about incomplete, imprecise, or symbolic
knowledge. Rather than relying on precise numerical values, it focuses
on relational abstractions that capture essential spatial and temporal
configurations. It is particularly suited to commonsense reasoning
and to domains where metric information is unavailable or unnecessary~\cite{ligozat2013qualitative}.

Qualitative spatial and temporal reasoning builds on this idea through
formalisms such as Allen’s interval algebra~\cite{Allen1983} and
region-based calculi like RCC~\cite{randell1992spatial}, which provide
finite sets of qualitative relations equipped with algebraic operations
and constraint-based reasoning mechanisms. These frameworks have been
widely applied in areas including robotics, geographic information
systems, and activity recognition, and are supported by solid theoretical
foundations~\cite{ligozat2013qualitative}. Most existing calculi,
however, operate at a single level of granularity, whereas many real-world
problems inherently involve multiple scales, motivating the development
of multi-scale qualitative reasoning frameworks.

Many reasoning tasks in space and time inherently involve information
available at different levels of precision. Multi-scale formalisms~\cite{euzenat2001granularity,li2007qualitative}
are designed to represent and reason about such heterogeneous information
by organizing descriptions hierarchically across several levels of
granularity. In temporal reasoning, these levels may correspond to
minutes, hours, days, or weeks, while in spatial reasoning they may
range from fine metric units, such as centimeters, to coarser abstractions,
such as buildings, cities, or regions. Each scale provides a partial
and approximate view of reality, suited to a specific level of analysis.

Crucially, qualitative relations observed at one scale do not necessarily
carry over unchanged to another. A coarse description may mask distinctions
that become visible at a finer scale, while a fine-grained description
may be unnecessarily detailed or even unavailable at a higher level
of abstraction. For example, two objects may appear to be in contact
when observed from a distant or imprecise viewpoint, whereas a more
detailed observation reveals a small gap between them. Ignoring the
precision associated with such observations can therefore lead to
inconsistent or misleading conclusions.

Multi-scale qualitative reasoning aims to address this issue by explicitly
modeling the relationships between qualitative descriptions at different
granularities. By linking coarse and fine qualitative relations in
a principled way, multi-scale frameworks enable sound reasoning across
levels of abstraction, allowing information to be refined, approximated,
or reconciled as the scale changes. This capability is essential for
faithfully capturing real-world spatio-temporal phenomena and for
supporting robust reasoning in the presence of varying degrees of
precision.

However, existing approaches to multi-scale qualitative temporal reasoning
suffer from fundamental limitations. Classical multi-scale qualitative
frameworks, based on fixed quantitative granularities, are not complete:
they do not, in general, allow one to decide satisfiability. Moreover,
they are poorly suited to temporal reasoning in natural language,
where scales are typically implicit, vague, or context-dependent rather
than explicitly defined. To address this latter issue, qualitative
scales have been introduced, yielding a complete reasoning framework~\cite{cohen2015algebra},
that is, it allows satisfiability to be decided. Nevertheless, this
setting falls outside the scope of the multi-algebra framework~\cite{cohen2017checking,cohen2025deciding},
an algebraic formalism designed to encompass a wide range of combinations
of qualitative reasoning formalisms, including quantitative multi-scale
qualitative reasoning. Beyond providing a general setting for reasoning
and satisfiability checking, the multi-algebra framework has proved
particularly effective for identifying tractable fragments, thanks
to a number of general structural results established in this context~\cite{cohen2025deciding,cohenminimality,cohen2020tractable,cohen2017temporal}.
Since the multi-algebra framework does not apply in the context of
qualitative scales, its tools for identifying tractable classes therefore
cannot be used.

In this article, we propose, as a contribution, a third approach:
multi-scale reasoning based on non-discrete quantitative scales. As
we show, this framework enjoys both completeness and compatibility
with the multi-algebra setting, enabling the identification of tractable
fragments for multi-scale qualitative temporal reasoning. Furthermore,
we demonstrate that the tractability results obtained in this new
setting carry over to the framework of qualitative scales, thereby
highlighting the practical and theoretical relevance of our approach.
We then identify the largest tractable fragments of multi-scale temporal
reasoning on points (and, respectively, intervals) and show that the
maximal tractability of the classical case extends to the multi-scale
setting. Beyond temporal reasoning, our framework opens new perspectives
for analogical multi-scale reasoning, where scales locally provide
an unbounded amount of detail and can be continuously refined by zooming.
This includes reasoning about space and time grounded in human perception
or based on analog sensors, such as microscopes or telescopes.

The remainder of the paper is organized as follows. Section~\ref{sec:Background}
reviews background material on qualitative reasoning, multi-scale
reasoning, and the multi-algebra framework, together with its use
in multi-scale settings. Section~\ref{subsec:Mod=0000E8le-oo} introduces
our new multi-scale semantics based on non-discrete quantitative (analogical)
scales. In Section~\ref{subsec:Strong-Semantics}, we establish key
properties of this semantics, in particular that it forms a sequential
formalism to which the multi-algebra framework applies. Section~\ref{subsec:Compl=0000E9tude}
proves the completeness of the proposed framework for multi-scale
reasoning. Section~\ref{subsec:Transfert} shows that the tractability
results obtained in our setting are inherited by the framework of
qualitative scales. Section~\ref{subsec:Propri=0000E9t=0000E9s-des-op=0000E9rateurs}
studies the properties of the reasoning operators. Finally, Section~\ref{subsec:Classes-traitables}
exploits all previously established results to apply multi-algebra
tractability theorems and identify tractable fragments for multi-scale
qualitative temporal reasoning for our new semantics and thus for
the qualitative scales context.

\section{Background}\label{sec:Background}

In this section, we recall the foundations of qualitative reasoning,
in particular those based on point and interval algebras (Section~\ref{subsec:Qualitative-Reasoning}),
as well as the basics of multi-scale reasoning (Section~\ref{subsec:Multi-Scale-Qualitative-Reasonin}),
the multi-algebra framework (Section~\ref{subsec:Multi-algebra-Framework}),
and its application in the multi-scale context (Section~\ref{subsec:Multi-algebra-of-Multi-Scales}).

\subsection{Qualitative Reasoning}\label{subsec:Qualitative-Reasoning}

Two fundamental qualitative temporal algebras are the Point Algebra
($\PA$) and Allen’s Interval Algebra ($\IA$). The universe $\U_{\PA}=\T$
of $\PA$ is a set of temporal points, usually interpreted as instants
on a linearly ordered time domain, typically $\mathbb{R}$ or $\mathbb{Q}$,
whereas the universe $\U_{\IA}=\Int$ of $\IA$ is the set of temporal
intervals over a linearly ordered time domain. Thus, the entities
manipulated by these algebras are points and intervals, respectively.

Every relation of an algebra $\A$ can be decomposed into a unique
union of special relations, called basic relations. $\PA$ has three
basic relations: $<$ (before), $=$ (equal), and $>$ (after), whereas
$\IA$ has thirteen basic relations, described in Table~\ref{tab:def_bases_IC}.
Thus, any relation in $\PA$ or $\IA$ is a union of some of these
basic relations. The sets of basic relations are denoted $\Base_{\PA}$
and $\Base_{\IA}$, the universal relations (i.e. the union of all
basic relations) by $\B_{\PA}$ and $\B_{\IA}$, and the empty relations
by $\vide$.
\begin{table}
\begin{centering}
\begin{tabular}{|c|c|c|}
\hline 
\multicolumn{2}{|c|}{Basic relations} & Definition\tabularnewline
\hline 
\hline 
$\contrainte x{\text{before}}y$ & $\correctTable$$\contrainte y{\text{after}}x$ & $x^{-}<x^{+}<y^{-}<y^{+}$\tabularnewline
\hline 
$\contrainte x{{\rm\text{meets}}}y$ & $\correctTable$$\contrainte y{\text{met by}}x$ & $x^{-}<x^{+}=y^{-}<y^{+}$\tabularnewline
\hline 
$\contrainte x{\text{overlaps}}y$ & $\correctTable$$\contrainte y{\text{overlapped by}}x$ & $x^{-}<y^{-}<x^{+}<y^{+}$\tabularnewline
\hline 
$\contrainte x{\text{starts}}y$ & $\correctTable$$\contrainte y{\text{started by}}x$ & $x^{-}=y^{-}<x^{+}<y^{+}$\tabularnewline
\hline 
$\contrainte x{\text{during}}y$ & $\correctTable$$\contrainte y{\text{contains}}x$ & $y^{-}<x^{-}<x^{+}<y^{+}$\tabularnewline
\hline 
$\contrainte x{\text{finishes}}y$ & $\correctTable$$\contrainte y{\text{finished by}}x$ & $y^{-}<x^{-}<x^{+}=y^{+}$\tabularnewline
\hline 
\multicolumn{2}{|c|}{$\correctTable$$\contrainte x{\text{equals}}y$} & $x^{-}=y^{-}<x^{+}=y^{+}$\tabularnewline
\hline 
\end{tabular}
\par\end{centering}
\caption{$13$ Allen basic relations}\label{tab:def_bases_IC}
\end{table}

The relation operators of both algebras are the union $\cup$, the
intersection $\cap$, the converse $\inv{\cdot}$, and the composition
$\comp$. These operators allow standard qualitative inferences, such
as $\contrainte xry\implies\contrainte y{\inv r}x$, $\contrainte xry\et\contrainte x{r\lyxmathsym{’}}y\implies\contrainte x{(r\cap r\lyxmathsym{’})}y$,
and $\contrainte xry\et\contrainte y{r\lyxmathsym{’}}z\implies\contrainte x{(r\comp r\lyxmathsym{’})}z$,
with $r,r\lyxmathsym{’}\in\A$ and $x,y,z\in\U$ (i.e. being temporal
variables). The composition of $\PA$ basic relations is described
in Table~\ref{tab:compo_PC}. See~\cite{Allen1983} for the composition
table of Allen interval of basic relations. 
\begin{table}
\centering{}%
\begin{tabular}{|c|c|c|c|}
\hline 
$\comp$ & $<$ & $=$ & $>$\tabularnewline
\hline 
\hline 
$<$ & $<$ & $<$ & $\B_{\PA}$\tabularnewline
\hline 
$=$ & $<$ & $=$ & $>$\tabularnewline
\hline 
$>$ & $\B_{\PA}$ & $>$ & $>$\tabularnewline
\hline 
\end{tabular}\caption{Composition table of $\protect\PA$ basic relations}\label{tab:compo_PC}
\end{table}

Descriptions based on $\PA$ or $\IA$ are formalised as qualitative
constraint networks, i.e., conjunctions of relations between temporal
variables. Satisfiability for $\PA$ networks is decidable in polynomial
time by enforcing algebraic closure (path-consistency), whereas for
general $\IA$ networks it is $\NP$-complete. Tractable subsets of
$\IA$ have been identified, such as the preconvex subclass $\IAp$~\cite{ligozat1994tractable,nebel1995reasoning,krokhin2003reasoning}
(the largest maximal tractable subclass of $\IA$), allowing polynomial-time
reasoning. A formal definition of a qualitative formalism is provided
in the following definition:
\begin{defn}
\cite{cohen2025deciding}A \emph{symmetric qualitative formalism}
is a triplet $(\Aa,\U,\I)$, where $(\Aa,\sunion,\scompl{\cdot},\svide,\B,\circ,\sinv{\cdot},\se)$
forms a finite non-associative algebra~\cite{ligozat2004qualitative},
$\U$ is a non-empty set, the entity domain, and $\I$ is a function
$\I\colon\Aa\to2^{\U\times\U}$, the interpretation function, that
satisfies, for all $r,r'\in\A$, with $x\sinter y=\scompl{(\scompl x\sunion\scompl y)}$,
the following properties: 
\begin{align*}
\begin{gathered}\I(\svide)=\vide\\
\I(\sinv r)=\inv{\I(r)}
\end{gathered}
 &  & \begin{gathered}\I(r\sinter r')=\I(r)\cap\I(r')\\
\I(r\sunion r')=\I(r)\cup\I(r')\\
\I(r\circ r')\supseteq\bigl(\I(r)\comp\I(r')\bigr)\cap\I(\B)
\end{gathered}
\end{align*}
\end{defn}

\subsection{Multi-Scale Qualitative Reasoning }\label{subsec:Multi-Scale-Qualitative-Reasonin}

In this section, we recall the foundations of multi-scale qualitative
temporal reasoning.

In this context, time is represented by a temporal domain $\mathcal{T}$,
which is a totally ordered dense or discrete set. Its elements are
called temporal points or instants. Temporal scales are then discretizations
of time with different resolutions.
\begin{defn}
\cite{Bettini2000,euzenat2005time}\label{def:granularity} A temporal
scale (or granularity) $g$ over $\mathcal{T}$ is a function from
a totally ordered set $I_{g}$ to the power set of the temporal domain,
$2^{\mathcal{T}}$, such that:

\[
\forall i,j,k\in I_{g}\sep i<k<j\et g(i)\neq\vide\et g(j)\neq\vide\implies g(k)\neq\vide
\]
\[
\forall i,j\in I_{g}\sep i<j\implies\forall x\in g(i)\minisep\forall y\in g(j)\sep x<y
\]
\end{defn}

Thus, each scale is a family of subsets of the temporal domain that
preserves the natural ordering of instants. The sets $g(i)$ are called
the granules of the scale $g$; they are the “instants” of $g$.
\begin{example}
The Gregorian calendar is a set of scales. At the day scale, the granule
is the day. Consequently, at this scale, all events occurring on the
same day are considered to occur at the same time.
\end{example}

Note that temporal scales are required to be discrete.~\cite{Bettini2000,euzenat2005time},
which is not necessarily the case in this article. Note also that
other formalizations of temporal scales, based on logic, have been
proposed.~\cite{montanari1996metric,franceschet2001dividing,euzenat2005time}.

Scales are related to each other through particular relations, namely
partial orders, that capture the intuitive notion of being “more precise”
or having a higher “resolution”:
\begin{defn}
\cite{Bettini2000}.\label{def:relations_entre_echelles}Let $g$
and $h$ be two scales over $\T$. 
\begin{itemize}
\item $g$ is finer\emph{ }(or less coarse) than $h$, denoted by $g\preceq h$,
if and only if $\forall i\in I_{g}\minisep\exists j\in I_{h}\sep g(i)\subseteq h(j)$,
that is, each granule of $g$ is contained within a granule of $h$.
\item $g$ \emph{decompose} $h$, denoted by $g\trianglelefteq h$ if and
only if $\forall j\in I_{h}\minisep\exists S\subseteq I_{g}\sep h(j)=\bigcup_{i\in S}g(i)$,
that is, each granule of $h$ is composed of granules of $g$.
\end{itemize}
\end{defn}

\begin{example}
Consider the Gregorian calendar. The scale of months is finer than
that of years. However, the scale of weeks is not finer than that
of months (and vice versa), since any period of time expressed in
months cannot be expressed as a set of weeks.
\end{example}

A scale $g$ maps a point $x$ to its \emph{representative} $\repr\left(x\right)$
at this scale which corresponds to the index $i\in I_{g}$ such that
$x\in g(i)$. From this modeling of time, we can represent information
defined at different scales: two entities satisfy a constraint \emph{defined
at a scale} if the representatives of the two entities at this scale
satisfy the corresponding ``classical'' constraint. For example,
having $x<y$ at the minute scale means that the minute of $x$ is
before the minute in which $y$ occurs.

Conversion operators have been introduced to reason on multiple scales
to convert information from one scale to another~\cite{euzenat2005time}.
In the qualitative framework, the \emph{downward conversion} $\downgc$
allows propagating relations to a finer scale and \emph{upward conversion}
$\upgc$ allows propagating relations to a \emph{coarser} scale. Table~\ref{tab:conversions_PA}
describes the conversions of the point algebra. For example, since
one instant may be before another at the scale of seconds while taking
place during the same minute, the upward conversion verifies $\{=\}\in\upgc\{<\}$.
More precisely, $\upgc\{<\}=\{<,=\}$. Thus, if we know that $x<y$
at the scale of seconds, we can deduce that $x\leq y$ at the scale
of minutes. Table 7 of \cite{euzenat1995space} describes the upward
and downward conversion for $\IA$.

A formalization of qualitative multiscale temporal reasoning over
points and interval with qualitative scales have been proposed~\cite{cohen2015algebra}.
In this context, when one seeks to decide satisfiability, the question
is whether there exist both temporal entities (points/intervals) and
discrete scales that satisfy the multi-scale constraints (for example,
$x<y$ at scale $g_{0}$ and $x=y$ at scale $g_{1}$), while also
satisfying the inter-scale constraints (for example, $g_{0}\preceq g_{1}$).
In other words, in the context of qualitative scales, the discrete
scales are also variables, subject to additional constraints between
them, related to their relative precision.

\begin{table}
\centering{}%
\begin{tabular}{cccc}
\toprule 
$b$ & $<$ & $=$ & $>$\tabularnewline
\midrule 
$\upgc b$ & $\leq$ & $=$ & $\geq$\tabularnewline
$\downgc b$ & $<$ & $\mathcal{B}_{\PA}$ & $>$\tabularnewline
\end{tabular}\caption{Upward and downward conversions of the point algebra}\label{tab:conversions_PA}
\end{table}

\subsection{Multi-algebra Framework}\label{subsec:Multi-algebra-Framework}

In this section, we recall the foundations of the multi-algebra framework.

A multi-algebra\cite{cohen2025deciding} is a Cartesian product of
non-associative algebras $\mathcal{A}_{0}\times\cdots\times\mathcal{A}_{m-1}$,
endowed with projections, that is, functions from $\mathcal{A}_{i}$
to $\mathcal{A}_{j}$ for distinct $i,j\in\J$, satisfying specific
properties. They define the structure of a form of combination of
qualitative formalisms. A relation in this context is therefore a
sequence of classical relations. In the multi-scale setting, each
element of the sequence corresponds to a different scale.

When equipped with an appropriate semantics, multi-algebras constitute
sequential qualitative formalisms, for which tractability results
have been identified \cite{cohen2025deciding}:
\begin{defn}
\cite{cohen2025deciding}A \emph{(qualitative) sequential formalism}
is a triple $(\A,\U,\I)$, where $\A$ is a finite multi-algebra,
$\U$ is a non-empty set, the \emph{universe}, and $\I$ is a function
$\I\colon\A\to2^{\U\times\U}$, the \emph{interpretation function},
satisfying, for all $R,R'\in\A$, the following properties:
\begin{align*}
\begin{gathered}\begin{aligned}\I(\cconv R) & =\I(R)\\
\I(\sinv R) & =\overline{\I(R)}
\end{aligned}
\\
\I\bigl(\ctrois{\svide}{\ldots}{\svide}\bigr)=\vide
\end{gathered}
 &  & \begin{aligned}\I(R\circ R') & \supseteq\bigl(\I(R)\circ\I(R')\bigr)\cap\I\left(\B\right)\\
\I(R\sinter R') & =\I(R)\cap\I(R')\\
\I(R) & =\bigcup_{B\atome R}\I(B)\text{.}
\end{aligned}
\end{align*}
\end{defn}

A large family of combinations of qualitative formalisms is loose
integration~\cite{wolfl2009combinations}, which constitutes a type
of sequential qualitative formalism~\cite{cohen2025deciding}.
\begin{defn}
\cite{cohen2025deciding}\emph{The loose integration} of $m$ symmetric
qualitative formalisms on the same universe $\U$, $(\mathcal{\A}_{0},\U,\I_{0}),\dots,(\mathcal{\A}_{m-1},\U,\I_{m-1})$,
is the triple $(\A,\U,\I)$, where $\A$ is the multi-algebra whose
Cartesian product is $\A_{0}\times\dots\times\A_{m-1}$ verifying
the following property, for all distinct $i,j\in\J$ and $b\in\Base{}_{i}$
: 
\[
\conv ijb=\sbigunion\liste{b'\in\Base_{j}}{\I_{i}(b)\cap\I_{j}(b')\neq\vide}
\]
and where $\I$ is defined by $\forall R\in\A\sep\I(R)=\bigcap_{i}\I_{i}(R_{i})$.
\end{defn}

In the context of multi-algebras, descriptions are \emph{(qualitative
constraint) network}s. A network over a multi-algebra $\A$ is a set
of variables $\E$ and a function $N$ associating with each pair
of variables $(x,y)\in\E^{2}$ such that $x\neq y$ a relation of
$\A$ and satisfying $N(x,y)=\inv{N(y,x)}$ for all distinct $x,y\in\E$.
A single constraint network whose relations are sequences of relations,
i.e. relations of a multi-algebra can thus represent a sequence of
classical constraint networks. The relation $N(x,y)$ is denoted more
succinctly by $N^{xy}$. Sometimes, it is useful to refer to the ``subnetwork''
associated with the index $i\in\I$ of a network $N$. This subnetwork
is denoted $N_{i}$ and is defined by $\left(N_{i}\right)^{xy}=\left(N^{xy}\right)_{i}$
for all distinct $x,y\in\E$. When $N$ is a multi-scale description,
$N_{i}$ describes the relations at the granularity $g_{i}$.Similarly,
the set $\liste{R_{i}}{R\in\S}$, with $\S\subseteq\A$, is denoted
$\S_{i}$, and corresponds to the subset of relations in $N_{i}$.
A network $N$ is \emph{over} a relation subset $\S\subseteq\A$ if
for all distinct $x,y\in\E$, $N^{xy}\in\S$. A \emph{scenario} is
a network over $\Base$.

A relation $R$ \emph{refines} a relation $R'$ if $R_{i}\subseteq R'_{i}$
for all $i\in\I$. More generally, $N$ \emph{refines} $N'$, which
is denoted $N\subseteq N'$, if for all distinct $x,y\in\E$, $N^{xy}\subseteq\left(N'\right)^{xy}$.
We denote $B\in R$ if and only if $B\subseteq R$ and $B$ is basic. 

A network $N$ is said \emph{$\comp$-closed} if it is closed under
composition, i.e. for all distinct $x,y,z\in\E$, $N^{xz}\subseteq N^{xy}\comp N^{yz}$.
A network $N$ is called \emph{algebraically closed} if it is \emph{$\comp$-closed},
and if each of its relations $N^{xy}$ is closed under projection,
i.e. for all distinct $i,j\in\I$, $N_{j}^{xy}\subseteq\conv{}{}_{i}^{j}N_{i}^{xy}$. 

From the semantics of sequential formalisms follows the definition
of satisfiability for networks, namely that the relations of the constraint
network are satisfied: 
\begin{defn}
\label{def:coherence_multi-reseau}Let $\F=(\A,\U,\I)$ be a sequential
formalism and $N$ be a network over $\A$.

A family $\{u_{x}\}_{x\in\mathtt{E}}$ is called \emph{solution} of
$N$ (for $\F$) if it satisfies $(u_{x},u_{y})\in\I(N^{xy})$, for
all distinct $x,y\in\E$.

The network $N$ is said \emph{satisfiable} (or \emph{consistent})
(for $\F$) if there exists a solution of $N$ (for $\F$) .
\end{defn}

A relation $R$ is \emph{trivially unsatisfiable} if there exists
$i\in\I$ such that $R_{i}=\vide$. A network is \emph{trivially
unsatisfiable }when there exists distinct $x,y\in\E$ such that $N^{xy}$
is a trivially unsatisfiable relation. A network $N$ is said \emph{$\comp$-consistent},
if it is \emph{$\comp$}-closed and not trivially inconsistent. In
the literature, \emph{$\comp$}-consistency is often conflated with
\emph{path-consistency}, because for some formalisms they are equivalent~\cite{renz2005weak}.
An algebraically closed network that is not trivially unsatisfiable
is said to be \emph{algebraically consistent}.

By restricting networks to some relation subsets $\S$, we get the
fundamental property: if the algebraic closure of a network over $\S$
is algebraically consistent, then this network is satisfiable. Such
subsets are \emph{algebraically tractable}: with an algebraically
tractable subset $S$, deciding the satisfiability of a network over
$S$ only requires to verify that its \textit{algebraic closure} is
not trivially inconsistent~\cite{cohen2025deciding}. A subset
$\S\subseteq\A$ is a \emph{subclass} when it is closed under intersection,
converse, and composition (i.e. for all $R,R'\in\S$, we have $R\cap R'\in\S$,
$\inv R\in\S$, and $R\comp R'\in\S$).\emph{ }Subsets containing
all the basic relations ($\S$ such that $\Base\subseteq\S$) are
said \emph{basic}. Basic subclasses are also called \emph{subalgebras}.
We say that a subset $\S\subseteq\A$ is \emph{$\conv{}{}$-closed}
when for all $R\in\S$, $\conv{}{}\left(R\right)\in\S$.

\subsection{Multi-algebra of Multi-Scales}\label{subsec:Multi-algebra-of-Multi-Scales}

We now turn to multi-algebras for multi-scale qualitative reasoning.

We present the multi-algebra, denoted $\SPA^{m}$~\cite{cohen2025deciding},
representing multi-scale temporal descriptions based on points, with
$m$ the number of scales. $\PA^{m}$ is its cartesian product. Scales
are considered totally ordered by the finesse relation $\preceq$.
In this case, the $i$-th $\PA$ corresponds to the $i$-th scale,
thus $R_{i}$ is the relation at the scale $g_{i}$. For example,
the relation $\cquatre ={\mathord{\mathord{<}\cup\mathord{=}}}{\mathord{\mathord{<}\cup\mathord{=}}}<$
of $\PA^{4}$ describes a possible temporal relation between two instants
at four different scales (days, hours, minutes, and seconds, for example).
This relation means that the points take place during the same day,
the first instant is not after the second at the scale of minutes
and at the scale of hours, and that the first instant is before the
second at the scale of seconds. The projections of $\SPA^{m}$ correspond
to the upward conversions $\upgc$ and the downward conversion $\downgc$.
More precisely, for all $b\in\Base_{\PA}$, we have $\conv ijb=\upgc b$
and $\conv jib=\downgc b$ if the scale $i$ is finer than the scale
$j$ (i.e. if $g_{i}\preceq g_{j})$. The multi-algebra for representing
multi-scale temporal descriptions based on intervals are analogous
(the cartesian product is $\IA^{m}$).

The interpretation of a multi-algebra relation in the context of quantitative
discrete scales is defined as the intersection of the interpretations
of the relations defined at each scale. Formally: for $R\in\PA^{m}$,
$\I(R)=\bigcup_{j\in\J}\I_{j}(R_{j})$ with $\varphi_{j}(r)=\liste{(x,y)\in\mathbb{\T}^{2}}{(g_{j}\left(x\right),g_{j}\left(y\right))\in\I_{\PA}(r)}$,
$g_{k}\left(x\right)$ the representative of the instant $x$ at the
scale $g_{k}$ and $\I_{\PA}(r)$ the interpretation function of the
point calculus (the classic qualitative formalism associated with
$\PA$). Note that such a formalism constitutes a sequential qualitative
formalism~\cite{cohen2025deciding}.

Note that the multi-algebra framework cannot be applied directly to
identify tractable subclasses in the context of qualitative scales.
These do not constitute a sequential formalism. Unlike the framework
of discrete quantitative scales, in this other context, qualitative
scales are additional variables, and that the base variables are linked
to them. In fact, one would need to generalize the multi-algebra framework
to include, alongside binary relations, ternary relations that coexist
with binary ones. We show in this article that it is not necessary
to introduce such a hypothetical framework in this context to obtain
the desired results.

\section{Contributions}\label{sec:Contributions}

\subsection{Our New Semantics: Analog Scales}\label{subsec:Mod=0000E8le-oo}

In this section, we propose a new semantics for the framework of multi-scale
point-based reasoning and multi-scale interval-based reasoning. It
is based on a set of rational scales, where the perception of time
is dense at each scale. These scales are totally ordered with respect
to the notion of scale precision. Each granule of a scale is composed
of an infinity of granules of any strictly finer scale. This amounts
to performing a zoom operation that adds infinitely many details.
This is a weaker constraint than the classical framework of quantitative
discrete scales, which adds only a finite and fixed number of details.
It also contrasts with the framework of qualitative discrete scales,
which add a finite but unconstrained number of details. 

We do not know whether these analogical scales are realistic models
of time, nor whether they can be used as a valid approximation of
time or space. If this turns out to be the case, using rational rather
than discrete scales could have potential applications in the context
of human perception of time and/or space, or in the context of analog
tools (optical devices, etc.). In these settings, discrete modeling
is possible but not optimal. Even if they are not a valid approximation,
we show that they are still useful, since they can serve as a proof
tool to establish certain properties of the qualitative scale framework.

Our new semantics is based on a mathematical property: the existence
of an order isomorphism for the lexicographic order between $\Q$
and $\Q\times\cdots\times\Q$. This isomorphism enables the multi-scale
encoding of time points and allows one to move from one scale to another.
Note that, there are infinitely many such isomorphisms. Let $\i_{j}$
be an isomorphism from $\Q$ to $\Q^{j}$. This isomorphism is fixed
throughout the article (its choice is not important for this article:
only order-theoretic properties of these isomorphisms are used in
the sequel; the results do not depend on any particular choice). In
the case of $\i_{1}$, the chosen isomorphism is the identity.
\begin{defn}
Let $j\in\J$ , we define the function $\gran_{j}$ from $I_{\gran_{j}}=\Q$
to $2^{\T}=2^{\Q}$ by associating each granular instant $i\in I_{\gran_{j}}$
with the granule $\gran_{j}\left(i\right)=\liste{t\in\Q}{\i_{m}\left(t\right)_{0,\ldots,j}=\i_{j+1}\left(i\right)}$.

The representative of the (non-granular) instant $t\in\T=\Q$ associated
with $\gran_{j}$, denoted $\repr_{j}\left(t\right)$, is defined
by $\repr_{j}\left(t\right)=\i_{j+1}^{-1}\left(\i_{m}\left(t\right)_{0,\ldots,j}\right)$.

The representative of the interval $[t,t']$ with $t,t'\in\T=\Q$
associated with $\gran_{j}$, denoted $\repr_{j}\left([t,t']\right)$,
is defined by $[\repr_{j}\left(t\right),\repr_{j}\left(t'\right)]$.
\end{defn}

Let us note that the objects $g_{j}$ thus defined are indeed temporal
scales and that they are totally ordered:
\begin{prop}
Let $j\in\J$ and $k\in\left\{ 0,\ldots,j-1\right\} $.

$g_{j}$ is a scale over $\Q$.

We have $g_{j}\preceq g_{k}$ and $g_{j}\trianglelefteq g_{k}$.
\end{prop}

\begin{proof}
We show that $g_{j}$ is a scale. We have the first property of the
definition of being a scale. Indeed, for all $i\in I_{g_{j}}=\Q$,
$\gran_{j}\left(i\right)\neq\vide$. We now show the second property
of scales. Let $i,k\in I_{g}$ such that $i<k$. Let $x\in\gran_{j}\left(i\right)$
and $y\in\gran_{j}\left(k\right)$. By definition, we have $x\in\liste{t\in\Q}{\i_{m}\left(t\right)_{0,\ldots,j}=\i_{j+1}\left(i\right)}$
and $y\in\liste{t\in\Q}{\i_{m}\left(t\right)_{0,\ldots,j}=\i_{j+1}\left(k\right)}$.
Thus, $\i_{m}\left(x\right)_{0,\ldots,j}=\i_{j+1}\left(i\right)$
and $\i_{m}\left(y\right)_{0,\ldots,j}=\i_{j+1}\left(k\right)$. However,
since $i<k$, we have $\i_{j+1}\left(i\right)<\i_{j+1}\left(k\right)$
and therefore $\i_{m}\left(x\right)_{0,\ldots,j}<\i_{m}\left(y\right)_{0,\ldots,j}$.
Consequently, $x<y$.

We now show that $g_{j}\preceq g_{k}$. Let $i\in I_{g_{j}}$. Therefore,
we have $i=\i_{j+1}^{-1}\left(\i_{m}\left(x\right)_{0,\ldots,j}\right)$.
Consider $l=\i_{k+1}^{-1}\left(\i_{m}\left(x\right)_{0,\ldots,k}\right)$.
We have $l\in I_{g_{k}}$. Let $x'\in\gran_{j}\left(i\right)$. We
just need to show that $x'\in\gran_{k}\left(l\right)$. Since $x'\in\gran_{j}\left(i\right)$,
we have $\i_{m}\left(x'\right)_{0,\ldots,j}=\i_{j+1}\left(i\right)=\i_{m}\left(x\right)_{0,\ldots,j}$.
Thus, $\i_{m}\left(x'\right)_{0,\ldots,k}=\i_{m}\left(x\right)_{0,\ldots,k}$
and therefore $\i_{k+1}^{-1}\left(\i_{m}\left(x'\right)_{0,\ldots,k}\right)=\i_{k+1}^{-1}\left(\i_{m}\left(x\right)_{0,\ldots,k}\right)$.
Consequently, $\i_{k+1}^{-1}\left(\i_{m}\left(x'\right)_{0,\ldots,k}\right)=l$.
In other words, $\i_{m}\left(x'\right)_{0,\ldots,k}=\i_{k+1}\left(l\right)$.
Thus, $x'\in\gran_{k}\left(l\right)$. 

We now show that $g_{j}\trianglelefteq g_{k}$. Let $l\in I_{g_{k}}$.
$\gran_{k}\left(l\right)=\liste{t\in\Q}{\i_{m}\left(t\right)_{0,\ldots,k}=\i_{k+1}\left(l\right)}$.

Consider $S\subseteq I_{g_{j}}$ defined by $S=\liste{\repr_{j}\left(x\right)}{x\in\gran_{k}\left(l\right)}$.
Thus, $S=\liste{\i_{j+1}^{-1}\left(\i_{m}\left(x\right)_{0,\ldots,j}\right)}{x\in\gran_{k}\left(l\right)}$.
However, $\bigcup_{i\in S}\gran_{j}\left(i\right)$ is equal to $\bigcup_{i\in S}\liste{t\in\Q}{\i_{m}\left(t\right)_{0,\ldots,j}=\i_{j+1}\left(i\right)}$,
that is $\bigcup_{i\in S}\liste{t\in\Q}{\i_{m}\left(t\right)_{0,\ldots,j}=\i_{j+1}\left(i\right)}$.
In other words, it is equal to $\liste{t\in\Q}{\exists x\in\gran_{k}\left(l\right)\et\i_{m}\left(t\right)_{0,\ldots,j}=\i_{j+1}\left(\i_{j+1}^{-1}\left(\i_{m}\left(x\right)_{0,\ldots,j}\right)\right)}$,
i.e. $\liste{t\in\Q}{\exists x\in\gran_{k}\left(l\right)\et\i_{m}\left(t\right)_{0,\ldots,j}=\i_{m}\left(x\right)_{0,\ldots,j}}$,
thus $\liste{t\in\Q}{\exists x\in\Q\et\i_{m}\left(x\right)_{0,\ldots,k}=\i_{k+1}\left(l\right)\et\i_{m}\left(t\right)_{0,\ldots,j}=\i_{m}\left(x\right)_{0,\ldots,j}}$.
Therefore, it is equal to $\liste{t\in\Q}{\i_{m}\left(t\right)_{0,\ldots,k}=\i_{k+1}\left(l\right)}$.
We thus have $\gran_{k}\left(l\right)=\bigcup_{i\in S}\gran_{j}\left(i\right)$.
\end{proof}
We can now define the interpretation of $\PA$ and $\IA$ relations
defined at one of our new scales.
\begin{defn}
Let $j\in\J$ and let $r\in\PA$, the interpretation of $r$ at scale
$\gran_{j}$ is defined by the function $\varphi_{\rgpc}^{j}\left(r\right)=\left\{ \left(x,y\right)\in\Q^{2}\ |\ \left(\repr_{j}\left(x\right),\repr_{j}\left(y\right)\right)\in\varphi_{\pc}\left(r\right)\right\} $
where $\varphi_{\pc}\left(r\right)$ is the classical interpretation
function of the point algebra.

Let $j\in\J$ and let $r\in\IA$, the interpretation of $r$ at scale
$\gran_{j}$ is defined by the function $\varphi_{\rgic}^{j}\left(r\right)=\left\{ \left(x,y\right)\in\Int^{2}\ |\ \left(\repr_{j}\left(x\right),\repr_{j}\left(y\right)\right)\in\varphi_{\IA}\left(r\right)\right\} $
where $\varphi_{\IA}\left(r\right)$ is the classical interpretation
function of the interval algebra and $\Int$ the intervals set of
$\Q$.
\end{defn}

Note that the properties of discrete scales are recovered in the context
of the new scales.
\begin{prop}
\label{prop:conversions}Let $x,y\in\Q$ and $j\in\J$. Let $k\in\left\{ 0,\ldots,j-1\right\} $.
\begin{enumerate}
\item If $x<y$ then $\repr_{j}\left(x\right)<\repr_{j}\left(y\right)$.
\item If $x=y$ then $\repr_{j}\left(x\right)=\repr_{j}\left(y\right)$.
\item If $\repr_{j}\left(x\right)<\repr_{j}\left(y\right)$ then $x<y$.
\item If $\repr_{j}\left(x\right)<\repr_{j}\left(y\right)$ then $\repr_{k}\left(x\right)\leq\repr_{k}\left(y\right)$.
\item If $\repr_{j}\left(x\right)=\repr_{j}\left(y\right)$ then $\repr_{k}\left(x\right)=\repr_{k}\left(y\right)$.
\item If $\repr_{k}\left(x\right)<\repr_{k}\left(y\right)$ then $\repr_{j}\left(x\right)<\repr_{j}\left(y\right)$.
\end{enumerate}
\end{prop}

\begin{proof}
The proofs of properties 1, 2, and 3 are analogous to the following.

Let $x,y\in\Q$, $j\in\J$ and $k\in\left\{ 0,\ldots,j-1\right\} $.

We prove property 4; we assume that $\repr_{j}\left(x\right)<\repr_{j}\left(y\right)$.
Thus $\i_{j+1}^{-1}\left(\i_{m}\left(t\right)_{0,\ldots,j}\right)<\i_{j+1}^{-1}\left(\i_{m}\left(t\right)_{0,\ldots,j}\right)$
and therefore $\i_{m}\left(t\right)_{0,\ldots,j}<\i_{m}\left(t\right)_{0,\ldots,j}$.
Consequently, $\i_{m}\left(t\right)_{0,\ldots,k}\leq\i_{m}\left(t\right)_{0,\ldots,k}$.
Hence $\i_{k+1}^{-1}\left(\i_{m}\left(t\right)_{0,\ldots,k}\right)\leq\i_{k+1}^{-1}\left(\i_{m}\left(t\right)_{0,\ldots,k}\right)$,
that is, $\repr_{k}\left(x\right)\leq\repr_{k}\left(y\right)$.

We prove property 5; we assume that $\repr_{j}\left(x\right)=\repr_{j}\left(y\right)$.
Thus $\i_{j+1}^{-1}\left(\i_{m}\left(t\right)_{0,\ldots,j}\right)=\i_{j+1}^{-1}\left(\i_{m}\left(t\right)_{0,\ldots,j}\right)$
and therefore $\i_{m}\left(t\right)_{0,\ldots,j}=\i_{m}\left(t\right)_{0,\ldots,j}$.
Consequently, $\i_{m}\left(t\right)_{0,\ldots,k}=\i_{m}\left(t\right)_{0,\ldots,k}$.
Hence $\i_{k+1}^{-1}\left(\i_{m}\left(t\right)_{0,\ldots,k}\right)=\i_{k+1}^{-1}\left(\i_{m}\left(t\right)_{0,\ldots,k}\right)$,
that is, $\repr_{k}\left(x\right)=\repr_{k}\left(y\right)$.

We prove property 6; we assume that $\repr_{k}\left(x\right)<\repr_{k}\left(y\right)$.
Thus, $\i_{j+1}^{-1}\left(\i_{m}\left(t\right)_{0,\ldots,k}\right)<\i_{j+1}^{-1}\left(\i_{m}\left(t\right)_{0,\ldots,k}\right)$
and therefore $\i_{m}\left(t\right)_{0,\ldots,k}<\i_{m}\left(t\right)_{0,\ldots,k}$.
Consequently, $\i_{m}\left(t\right)_{0,\ldots,j}<\i_{m}\left(t\right)_{0,\ldots,j}$.
Hence $\i_{k+1}^{-1}\left(\i_{m}\left(t\right)_{0,\ldots,j}\right)<\i_{k+1}^{-1}\left(\i_{m}\left(t\right)_{0,\ldots,j}\right)$,
that is, $\repr_{j}\left(x\right)<\repr_{j}\left(y\right)$.
\end{proof}

\subsection{Multi-Algebra Semantics}\label{subsec:Strong-Semantics}

This section shows that the semantics introduced in the previous section,
namely the analog scales, induces a sequential qualitative formalism,
thereby making the full machinery of the multi-algebra framework applicable.

We begin by showing that reasoning with a single scale based on our
new semantics is a classical qualitative formalism.

\begin{prop}
Let $j\in\J$.
\begin{itemize}
\item $\left(\PA,\Q,\varphi_{\rgpc}^{j}\right)$ is a qualitative formalism.\label{prop:FQ}
\item $\left(\IA,\Int,\varphi_{\rgic}^{j}\right)$ is a qualitative formalism.
\end{itemize}
\end{prop}

\begin{proof}
Let $\left(\left(\mathcal{\A},\U,\I\right),\I'\right)\in\left\{ \left(\left(\PA,\Q,\varphi_{\rgpc}^{j}\right),\I_{\PC}\right),\left(\left(\IA,\Int,\varphi_{\rgic}^{j}\right),\I_{\IC}\right)\right\} $.
$\A$ is a non-associative algebra, $\I$ is a fonction from $\Aa$
to $2^{\U\times\U}$. It remains to verify that the properties of
$\I$ hold. Let $r,r'\in\A$. We trivially have $\I\left(\svide\right)=\svide$
and $\I(\sinv r)=\inv{\I(r)}$. We now show that $\I\left(r\right)\cup\I\left(r'\right)=\I\left(r\cup r'\right)$:

\[
\begin{array}{cc}
 & \I\left(r\right)\cup\I\left(r'\right)\\
= & \left\{ \left(x,y\right)\in\U^{2}\ |\ \left(\repr_{j}\left(x\right),\repr_{j}\left(y\right)\right)\in\varphi'\left(r\right)\right\} \\
 & \cup\\
 & \left\{ \left(x,y\right)\in\U^{2}\ |\ \left(\repr_{j}\left(x\right),\repr_{j}\left(y\right)\right)\in\varphi'\left(r'\right)\right\} \\
= & \left\{ \left(x,y\right)\in\U^{2}\ |\ \left(\repr_{j}\left(x\right),\repr_{j}\left(y\right)\right)\in\varphi'\left(r\right)\right.\\
 & \ou\\
 & \left.\left(\repr_{j}\left(x\right),\repr_{j}\left(y\right)\right)\in\varphi'\left(r'\right)\right\} \\
= & \left\{ \left(x,y\right)\in\U^{2}\ |\ \left(\repr_{j}\left(x\right),\repr_{j}\left(y\right)\right)\in\varphi'\left(r\right)\cup\varphi'\left(r'\right)\right\} \\
= & \left\{ \left(x,y\right)\in\U^{2}\ |\ \left(\repr_{j}\left(x\right),\repr_{j}\left(y\right)\right)\in\varphi'\left(r\cup r'\right)\right\} \\
= & \I\left(r\cup r'\right)
\end{array}
\]
The proof is analog for $(r\sinter r')=\I(r)\cap\I(r')$. Note that
$\I(\B)=\U\times\U=\Q\times\Q$. It remains to prove that $\I(r\comp r')\supseteq\bigl(\I(r)\comp\I(r')\bigr)$.
\[
\begin{array}{cc}
 & \I(r)\comp\I(r')\\
= & \liste{\left(x,y\right)\in\U^{2}}{\left(\repr_{j}\left(x\right),\repr_{j}\left(y\right)\right)\in\varphi'\left(r\right)}\\
 & \comp\\
 & \liste{\left(y,z\right)\in\U^{2}}{\left(\repr_{j}\left(y\right),\repr_{j}\left(z\right)\right)\in\varphi'\left(r'\right)}\\
= & \left\{ \left(x,z\right)\in\U^{2}\;|\ \exists y\in\U\et\left(\repr_{j}\left(x\right),\repr_{j}\left(y\right)\right)\in\varphi'\left(r\right)\right.\\
 & \et\\
 & \left.\left(\repr_{j}\left(y\right),\repr_{j}\left(z\right)\right)\in\varphi'\left(r'\right)\right\} \\
\subseteq & \left\{ \left(x,z\right)\in\U^{2}\;|\ \exists y_{g_{j}}\in\Repr_{j}\et\left(\repr_{j}\left(x\right),y_{g_{j}}\right)\in\varphi'\left(r\right)\right.\\
 & \et\\
 & \left.\left(y_{g_{j}},\repr_{j}\left(z\right)\right)\in\varphi'\left(r'\right)\right\} \\
= & \liste{\left(x,z\right)\in\U^{2}}{\left(\repr_{j}\left(x\right),\repr_{j}\left(z\right)\right)\in\varphi'\left(r\comp r'\right)}\\
= & \I(r\comp r')
\end{array}
\]
\end{proof}
Since the single-scale reasoning formalism is a qualitative formalism,
we can combine them using loose integration~\cite{cohen2025deciding}
to obtain a multi-scale formalism.
\begin{defn}
\label{def:combi_lache_raisonnement_rationnel_MS}$\rmgpc$ is the
loose integration of the qualitative formalisms $\left(\PA,\Q,\varphi_{\rgpc}^{j}\right)$
with $j\in\J$, that is the triplet $\rmgpc=\left(\PA^{m},\Q,\varphi_{\rgpc}\right)$
satisfying for all $R\in\PA^{m}$, $\varphi_{\rgpc}^ {}\left(R\right)=\bigcup_{j\in\J}\varphi_{\rgpc}^{j}R_{j}$
such that the projections of of the multi-algebra $\PA^{m}$ are defined
by the following proposition: $\conv ijb=\sbigunion\liste{b'\in\Base_{\PA}}{\varphi_{\rgpc}^{i}b)\cap\varphi_{\rgpc}^{j}(b')\neq\vide}$
for all $b\in\Base_{\PA}$.

$\rmgpc$ (resp. $\rmgic$) is the loose integration of the qualitative
formalisms $\left(\IA,\Int,\varphi_{\rgic}^{j}\right)$ with $j\in\J$,
that is the triplet $\rmgic=\left(\IA^{m},\Int,\varphi_{\rgic}\right)$
satisfying for all $R\in\IA^{m}$, $\varphi_{\rgic}\left(R\right)=\bigcup_{j\in\J}\varphi_{\rgic}^{j}R_{j}$
such that the projections of the multi-algebra $\IA^{m}$ are defined
by the following proposition: $\conv ijb=\sbigunion\liste{b'\in\Base_{\IA}}{\varphi_{\rgic}^{i}b)\cap\varphi_{\rgic}^{j}(b')\neq\vide}$
for all $b\in\Base_{\IA}$.
\end{defn}

We now show that the projections of our multi-scale formalisms are
indeed the upward and downward conversions.
\begin{prop}
Let $j\in\J$ and $k\in\left\{ 0,\ldots,j-1\right\} .$\label{prop:MA_de_FQ}

For all $r\in\PA$ (resp. all $r\in\IA)$, we have $\conv kjr=\downconv r$
and $\conv jkr=\upconv r$.
\end{prop}

\begin{proof}
To do this, we just need to show that the upward and downward conversions
correspond exactly to the conversion operators $\conv ij$ of the
multi-scale point (resp. interval) algebra defined by: 
\[
\conv ijb=\sbigunion\liste{b'\in\Base_{j}}{\I_{i}(b)\cap\I_{j}(b')\neq\vide}
\]
\begin{itemize}
\item On the one hand, if $i>j$, we show that $\conv ij\left\{ <\right\} =\sbigunion\liste{b'\in\Base_{j}}{\I_{i}(<)\cap\I_{j}(b')\neq\vide}=\left\{ <,=\right\} $.
\end{itemize}
Let $\left(x,y\right)\in\I_{i}(<)$. We have $\repr_{i}\left(x\right)<\repr_{i}(y)$
and therefore $\repr_{j}\left(x\right)\leq\repr_{j}(y)$ (Proposition~\ref{prop:conversions}).
In other words $\left\{ >\right\} \notin\conv ij\left\{ <\right\} $.
It remains to show that there exists $x,y\in\U$ such that $\repr_{i}\left(x\right)<\repr_{i}(y)$
and $\repr_{j}\left(x\right)<\repr_{j}(y)$ thus there exists $x',y'\in\U$
such that $\repr_{i}\left(x'\right)<\repr_{i}(y')$ and $\repr_{j}\left(x'\right)=\repr_{j}(y')$.
We take, for example $x=\i_{m}^{-1}\left(0,\ldots,0\right)$ and $y=\i_{m}^{-1}\left(1,\ldots,1\right)$
as well as $x'=\i_{m}^{-1}\left(0,\ldots,0\right)$ and $y'=\i_{m}^{-1}\left(\tilde{y}\right)$
with $\tilde{y}_{0,\ldots,j}=\left(0,\ldots,0\right)$ and $\tilde{y}_{j+1,\ldots,m-1}=\left(1,\ldots,1\right)$.
\begin{itemize}
\item On the other hand, if $i>j$, we show that $\conv ij\left\{ =\right\} =\sbigunion\liste{b'\in\Base_{j}}{\I_{i}(=)\cap\I_{j}(b')\neq\vide}=\left\{ =\right\} $.
\end{itemize}
Let $\left(x,y\right)\in\I_{i}(=)$. We have $\repr_{i}\left(x\right)=\repr_{i}(y)$
and therefore $\repr_{j}\left(x\right)=\repr_{j}(y)$ (Proposition~\ref{prop:conversions}).
In other words $\left\{ <,>\right\} \cap\conv ij\left\{ =\right\} =\vide$.
By taking, $x=y=\i_{m}^{-1}\left(0,\ldots,0\right)$, we have $\repr_{i}\left(x\right)=\repr_{i}(y)$
and $\repr_{j}\left(x\right)=\repr_{j}(y)$.
\begin{itemize}
\item On the one hand, if $i<j$, we show that $\conv ij\left\{ <\right\} =\sbigunion\liste{b'\in\Base_{j}}{\I_{i}(<)\cap\I_{j}(b')\neq\vide}=\left\{ <\right\} $.
\end{itemize}
Let $\left(x,y\right)\in\I_{i}(<)$. We have $\repr_{i}\left(x\right)<\repr_{i}(y)$
and therefore $\repr_{j}\left(x\right)<\repr_{j}(y)$ (Proposition~\ref{prop:conversions}).
In other words $\left\{ =,>\right\} \cap\conv ij\left\{ <\right\} =\vide$.
It remains to show that there exists $x,y\in\U$ such that $\repr_{i}\left(x\right)<\repr_{i}(y)$
and $\repr_{j}\left(x\right)<\repr_{j}(y)$. We take, for example
$x=\i_{m}^{-1}\left(0,\ldots,0\right)$ and $y=\i_{m}^{-1}\left(1,\ldots,1\right)$.
\begin{itemize}
\item On the other hand, if $i<j$, we show that $\conv ij\left\{ =\right\} =\sbigunion\liste{b'\in\Base_{j}}{\I_{i}(=)\cap\I_{j}(b')\neq\vide}=\left\{ <,=,>\right\} $.
\end{itemize}
Let's show that it exists $x,y\in\U$ such that $\repr_{i}\left(x\right)=\repr_{i}(y)$
and $\repr_{j}\left(x\right)<\repr_{j}(y)$, that it exists $x',y'\in\U$
such that $\repr_{i}\left(x'\right)=\repr_{i}(y')$ and $\repr_{j}\left(x'\right)=\repr_{j}(y')$,
and that it exists $x'',y''\in\U$ such that $\repr_{i}\left(x''\right)=\repr_{i}(y'')$
and $\repr_{j}\left(x'\right)>\repr_{j}(y')$. We take $x=\i_{m}^{-1}\left(0,\ldots,0\right)$
and $y=\i_{m}^{-1}\left(\tilde{y}\right)$ with $\tilde{y}_{0,\ldots,i}=\left(0,\ldots,0\right)$
and $\tilde{y}_{i+1,\ldots,m-1}=\left(1,\ldots,1\right)$, $x'=y'=x$,
and $x''=-x$, $y''=-y$.

The proof is analogous for the interval formalism.
\end{proof}
We conclude by showing that our new formalisms are indeed sequential
formalisms.
\begin{prop}
$\rmgpc$ (resp. $\rmgic$) is a sequential qualitative formalism.
\end{prop}

\begin{proof}
This follows directly from Definition~\ref{def:combi_lache_raisonnement_rationnel_MS},
 Proposition~\ref{prop:FQ} and Proposition 33 of~\cite{cohen2025deciding},
namely: any loose integration of qualitative formalism is a sequential
formalism.
\end{proof}

\subsection{Completeness}\label{subsec:Compl=0000E9tude}

In this section, we show that multi-scale reasoning based on our new
semantics is complete, that is, any algebraically closed description
using only basic relations is satisfiable. This is an essential property
for algebraic tractability.
\begin{thm}
$\rmgpc$ (resp. $\rmgic$) is complete.\label{thm:completude}
\end{thm}

\begin{proof}
The key idea of the proof is to combine consistent solutions at each
scale into a single global solution by exploiting the lexicographic
structure of the analog scales and its preservation by construction
using the isomorphisms $\i_{j}$. More precisely, given a consistent
solution for each scale $g_{j}$, we construct a global solution by
assigning to each variable a tuple in $\Q^{m}$ whose $j$-th component
corresponds to the solution at scale $g_{j}$. The lexicographic order
ensures that constraints satisfied at each scale are preserved in
the combined structure.

Let $N$ be an algebraically closed scenario over $\gpa$ (i.e. $N$
satisfies: $N^{x,y}\in\Base_{\PA}$ for all $x,y\in\E$). We therefore
just need to show that there exists a solution to $N$, that is entities
$\left\{ u_{x}\right\} _{x\in\E}\subseteq\Urgpc$ such that for all
distinct $x,y\in\E$ , we have $\left(u_{x},u_{y}\right)\in\varphi_{\rmgpc}\left(N^{x,z}\right)=\bigcap_{j\in\J}\varphi_{\rgpc}^{j}\left(N_{j}^{x,y}\right)$.
For this, we will construct a set of entities $\left\{ v_{x}\right\} _{x\in\E}\subseteq\Q^{m}$.

Let $j\in\J$. Consider $N_{j}$, which is associated with $g_{j}$.
$N_{j}$ has a solution in $\Q$ since $\pc$ is complete (in $\Q$),
let us denote it as $\left\{ u_{x}^{j}\right\} _{x\in\E}$. We then
define, for all $x\in\E$, $\left(v_{x}\right)_{j}=u_{x}^{j}$. Thus,
$\left\{ v_{x}\right\} _{x\in\E}$ is fully defined, and we indeed
have $\left\{ v_{x}\right\} _{x\in\E}\subseteq\Q^{m}$.

We now construct $\left\{ u_{x}\right\} _{x\in\E}$, : for all $x\in\E$,
$u_{x}=\i_{m}^{-1}\left(v_{x}\right)$. It therefore remains to show
that for all distinct $x,y\in\E$ and all $j\in\J$, we have $\left(u_{x},u_{y}\right)\in\varphi_{\rgpc}^{j}\left(N_{xy}^{j}\right)$.

To do this, let us first show that $\repr_{j}\left(u_{x}\right)=\i_{j+1}^{-1}\left(u_{x}^{0},\ldots,u_{x}^{j}\right)$.
By definition of $u_{x}$, $\repr_{j}\left(u_{x}\right)$ is the $i\in I_{g_{j}}=\Q$
such that $\i_{m}\left(u_{x}\right)_{0,\ldots,j}=i_{j+1}\left(i\right)$,
that is $\left(v_{x}\right)_{0,\ldots,j}=i_{j+1}\left(i\right)$.
Thus, by definition of $v_{x}$, $\left(u_{x}^{0},\ldots,u_{x}^{j}\right)=i_{j+1}\left(i\right)$
and therefore $\repr_{j}\left(u_{x}\right)=i=i_{j+1}^{-1}\left(u_{x}^{0},\ldots,u_{x}^{j}\right)$.

We now show that $\left(u_{x},u_{y}\right)\in\varphi_{\rgpc}^{j}\left(N_{xy}^{j}\right)$.
By construction, $\left(u_{x}^{j},u_{y}^{j}\right)\in\varphi_{\pc}\left(<\right)$
if and only if $\left(u_{x}^{0},\ldots,u_{x}^{j}\right)<\left(u_{y}^{0},\ldots,u_{y}^{j}\right)$
(indeed, since $N$ is closed under projection, we have the following
properties: if $N_{x,y}^{j}=\left\{ <\right\} $ then for all $k<j$,
either $N_{x,y}^{k}=\left\{ =\right\} $ or $N_{x,y}^{k}=\left\{ <\right\} $,
if $N_{x,y}^{j}=\left\{ =\right\} $ then for all $k<j$, $N_{x,y}^{k}=\left\{ =\right\} $,
and if there exists $k<j$ such that $N_{x,y}^{k}=\left\{ <\right\} $
then $N_{x,y}^{j}=\left\{ <\right\} $). Therefore $\left(u_{x}^{j},u_{y}^{j}\right)\in\varphi_{\pc}\left(<\right)$
if and only if $i_{j}^{-1}\left(u_{x}^{0},\ldots,u_{x}^{j}\right)<i_{j}^{-1}\left(u_{y}^{0},\ldots,u_{y}^{j}\right)$
(since $i_{j}^{-1}$ is an order isomorphism).

Thus, $\left(u_{x}^{j},u_{y}^{j}\right)\in\varphi_{\pc}\left(N_{x,y}^{j}\right)$
if and only if $\left(\i_{j}^{-1}\left(u_{x}^{0},\ldots,u_{x}^{j}\right),\i_{j}^{-1}\left(u_{y}^{0},\ldots,u_{y}^{j}\right)\right)\in\varphi_{\pc}\left(N_{xy}^{j}\right)$,
that is $\left(\repr_{j}\left(u_{x}\right),\repr_{j}\left(u_{y}\right)\right)\in\varphi_{\pc}\left(N_{xy}^{j}\right)$,
so $\left(u_{x},u_{y}\right)\in\varphi_{\rgpc}^{j}\left(N_{xy}^{j}\right)$
(definition of $\varphi_{\rgpc}^{j}$).

The proof is analogous for $\rmgic$.
\end{proof}

\subsection{Transfer}\label{subsec:Transfert}

In this section, we show that our new semantics enables the identification
of tractable subclasses within the qualitative scales framework by
leveraging the tractability theorems of the multi-algebra framework,
which cannot be directly applied to it since it is not a sequential
formalism.
\begin{thm}
Let $\S$ be an algebraically tractable $\conv{}{}$-closed subclass
of $\rmgpc$ or of $\rmgic$ . $\S$ is also algebraically tractable
for qualitative scales.\label{thm:transfert}
\end{thm}

\begin{proof}
Intuitively, this transfer is possible because qualitative scales
can be seen as a finite abstraction of the analog-scale semantics,
preserving satisfiability of algebraically closed scenarios. 

Let $\S$ be an algebraically tractable $\conv{}{}$-closed subclass
of $\rmgpc$ (resp. of $\rmgic$). Let $N$ be an algebraically consistent
network over $\rmgpc$. By algebraic tractability, it therefore contains
a satisfiable scenario, which is necessarily algebraically closed.
In the context of qualitative scales, algebraically closed scenarios
are satisfiable. Hence, $N$ is a satisfiable network for qualitative
scales. Consequently, for networks over $\S$ for qualitative scales,
algebraic consistency implies satisfiability. By Proposition 56, since
$\mathcal{S}$ is $\conv{}{}$-closed subclass, $\mathcal{S}$ is
algebraically tractable for qualitative scales.
\end{proof}

\subsection{Properties of Reasoning Operators}\label{subsec:Propri=0000E9t=0000E9s-des-op=0000E9rateurs}

In this section, we identify the properties of the reasoning operators
that will allow us, in the next section, to identify tractable classes.
\begin{prop}
Let $r,r'\in\PA$, we have:\label{prop:distributive}
\begin{itemize}
\item $\upconv r\comp\upconv r'=\upconv\left(r\comp r'\right)$,
\item $r\inter r'$ implies $\upconv\left(r\inter r'\right)=\upconv r\inter\upconv r'$.
\end{itemize}
Let $r,r'\in\IA$, we have:
\begin{itemize}
\item $\upconv r\comp\upconv r'\subseteq\upconv\left(r\comp r'\right)$,
\item $r\inter r'$ implies $\upconv\left(r\inter r'\right)=\upconv r\inter\upconv r'$.
\end{itemize}
\end{prop}

\begin{proof}
The underdistributivity of $\upconv$ is already known~\cite{euzenat1995time}.
The result follows straightforwardly by enumeration.
\end{proof}
\begin{prop}
Let $r\in\PA\cup\IA$.\label{prop:invariance}

We have $\upconv r=\upconv\hmax\left(r\right)$.
\end{prop}

\begin{proof}
The result follows straightforwardly by enumeration.
\end{proof}
\begin{prop}
We have:\label{prop:proj-clos}
\begin{itemize}
\item for all $r\in\IAp$, $\upconv r\in\IAp$,
\item for all $r\in\IAc$, $\upconv r\in\IAc$,
\item for all $r\in\PAc$, $\upconv r\in\PAc$. 
\end{itemize}
\end{prop}

\begin{proof}
The result follows straightforwardly by enumeration.
\end{proof}

\subsection{Tractable Fragments}\label{subsec:Classes-traitables}

In this section, we use the properties established in the previous
sections to apply the tractability theorems of the multi-algebra framework,
in order to identify the largest tractable subclass of multi-scale
point reasoning and multi-scale interval reasoning under our new semantics
and within the qualitative scales setting.

We begin by showing that the subclasses of convex relations for points
$\PAc$ and intervals $\IAc$~\cite{ligozat2013qualitative} are
tractable.
\begin{prop}
$\PAc^{m}$ and $\IAc^{m}$ are algebraically tractable subclasses.\label{prop:convexes_traitables}
\end{prop}

\begin{proof}
Let $\S\in\left\{ \PAc^{m},\IAc^{m}\right\} $. We apply the weakened
slicing theorem (Corollary 99 of \cite{cohen2017checking}) to prove
the first part of this lemma. We check its assumptions with the following
tree weakening: for all $i,j\in\left\{ 0,\ldots,m-1\right\} $ such
that $i<j$, for all $b\in\Base$, $\conv ijb=\B$:
\begin{itemize}
\item multi-algebras with totally ordered scales are tree multi-algebra: 
\begin{itemize}
\item the plenary anti-tree structure is the scales finest order: $0\rightarrow\cdots\rightarrow m-1$,
since $\upconv\upconv r=\upconv r$ and $\downconv\downconv r=\downconv r$~\cite{fisher2005handbook}.
\end{itemize}
\item algebraically closed scenarios are satisfiable (Theorem~\ref{thm:completude})
\item each $\S_{i}=$ is a basic subclass \cite{ligozat2013qualitative}.
\item each $\S_{i}=$ is minimal \cite{ligozat2013qualitative}.
\item $\S$ is $\conv{}{}$-distributive:
\begin{itemize}
\item for all $i,j\in\left\{ 1,\ldots,m\right\} $ such that $i<j$, $r\comp r'\neq\vide$
implies $\left(\conv ijr\right)\comp\left(\conv ijr'\right)\subseteq\left(\conv ijr\comp r'\right)$
and $r\inter r'\neq\vide$ implies $\left(\conv ijr\right)\inter\left(\conv ijr'\right)\subseteq\left(\conv ijr\inter r'\right)$
(Proposition~\ref{prop:distributive})
\item for all $i,j\in\left\{ 1,\ldots,m\right\} $ such that $i<j$, we
have $\left(\conv ijr\right)\comp\left(\conv ijr'\right)=\B=\left(\conv ijr\comp r'\right)$
and $\left(\conv ijr\right)\inter\left(\conv ijr'\right)=\B=\left(\conv ijr\inter r'\right)$.
\end{itemize}
\item for all distinct $i,j\in\J$, for all $r\in\S_{i}$, $\conv ijr\in\S_{j}$,
indeed:
\begin{itemize}
\item for all $i,j\in\left\{ 1,\ldots,m\right\} $ such that $i>j$, for
all $r\in\S$, $\conv ijr=\upconv r\in\S_{j}$ (Proposition~\ref{prop:proj-clos})
\item for all $i,j\in\left\{ 1,\ldots,m\right\} $ such that $i<j$, for
all $r\in\S$, $\conv ijr=\B\in\S_{j}$.
\end{itemize}
\end{itemize}
Thus $\S$ is algebraically tractable with and without the weakening.
\end{proof}
We now show that $\IAp$ the intervals subclasse of preconvex relations~\cite{ligozat2013qualitative}
and the full point algebra $\PA$ are tractable.

\begin{thm}
$\PA^{m}$ are $\IAp^{m}$ are algebraically tractable subclasses.\label{thm:application_heritage}
\end{thm}

\begin{proof}
Let $\S\in\left\{ \PA^{m},\IAp^{m}\right\} $. We apply the Weakened
Refinement Theorem (Corollary 100 of \cite{cohen2017checking}) by
checking its hypothesis, with the following weakening: for all $i,j\in\left\{ 0,\ldots,m-1\right\} $
such that $i>j$, for all $b\in\Base$, $\conv ijb=\B$.
\begin{itemize}
\item each $\S_{i}=$ is a subclass \cite{ligozat2013qualitative}.
\item each $\S_{i}=$ is composition stable through $\hmax$ \cite{renz1999maximal,ligozat2013qualitative}.
\item for all distinct $i,j\in\J$, $\S_{i}\times\S_{j}$ is projection
stable through $\hmax$:
\begin{itemize}
\item by Proposition 90 of \cite{cohen2025deciding} since each $\S_{i}$
is $\conv{}{}$-invariant through $\hmax$ (Proposition~\ref{prop:invariance}
for the case $i<j$ and since for all $b\in\Base$, $\conv ijb=\B$
for the case $i>j$)
\end{itemize}
\item for all distinct $i,j\in\J$, for all $r\in\S_{i}$, $\conv ijr\in\S_{j}$,
indeed:
\begin{itemize}
\item for all $i,j\in\left\{ 1,\ldots,m\right\} $ such that $i>j$, for
all $r\in\S$, $\conv ijr=\upconv r\in\S_{j}$ (Proposition~\ref{prop:proj-clos})
\item for all $i,j\in\left\{ 1,\ldots,m\right\} $ such that $i<j$, for
all $r\in\S$, $\conv ijr=\B\in\S_{j}$.
\end{itemize}
\item $\PAc^{m}$ and $\IAc^{m}$ are algebraically tractable (for the weakening
; proof of Proposition~\ref{prop:convexes_traitables}). 
\end{itemize}
\end{proof}
\begin{cor}
$\PA^{m}$ and $\IAp^{m}$ are algebraically tractable subclasses
within the setting of totally ordered qualitative scales.
\end{cor}

\begin{proof}
By Theorem~\ref{thm:application_heritage} and Theorem~\ref{thm:transfert}.
\end{proof}
Note that since $\PA$ and $\IAp$ are maximally tractable in the
classical setting, they are also maximally tractable in the multi-scale
context.
\begin{rem}
Note that these results also apply in the following alternative context,
having a different semantics: the two aligned scales: $g_{0}$ from
$\mathbb{Z}$ to $2^{\Q}$ (resp. $2^{\mathbb{R}}$) where $g_{0}\left(t\right)=\left[t,t+1\right[$
and $g_{1}$ from $\Q$ (resp. $\mathbb{R}$) to $2^{\Q}$ (resp.
$2^{\mathbb{R}}$) where $g_{1}\left(t\right)=\left\{ t\right\} $,
that is to say, to reason about continuous time and its natural discretization.
\end{rem}

\section{Conclusion}

We introduced a new semantics for multi-scale temporal reasoning based
on analog scales, which is both complete and compatible with the multi-algebra
framework.

This semantics induces a sequential qualitative formalism, allowing
the direct application of existing tractability results from the multi-algebra
setting. As a consequence, we identified the largest tractable subalgebra
of multi-scale temporal reasoning and showed that classical maximal
tractability results for the point algebra and the interval algebra
extend to the multi-scale setting. More precisely, this allowed us
to identify tractable fragments for our analog multi-scale reasoning
and to transfer these results to the qualitative scales setting. Note
that this is the first tractability result in the multi-scale qualitative
reasoning context.

The perspectives of this work include its generalization to the context
of multi-scale spatial reasoning, as well as to the more general multi-scale
temporal context, where intervals can become points when changing
scale. However, before undertaking this study, it is first necessary
to identify tractable subclasses in the context without scales of
reasoning about temporal entities that can be either points or intervals,
with the possibility of uncertainty about their type (interval or
point), by extending the study of tractability from the point-and-interval
context without uncertainty~\cite{jonsson2004complexity}.

\bibliographystyle{named}
\bibliography{biblio_fq}

\end{document}